\documentclass{article} %
\usepackage[preprint]{rssp_preprint}
\usepackage{xurl}
\usepackage{graphicx}
\usepackage{hyperref}
\usepackage{xcolor}
\definecolor{darkblue}{rgb}{0, 0, 0.5}
\hypersetup{colorlinks=true, citecolor=darkblue, linkcolor=darkblue, urlcolor=darkblue,
  pdftitle={Representation Redistribution in Sparse Autoencoders}, pdfauthor={Ziye Mei, Haotian Ma}}
\usepackage{natbib}
\usepackage{booktabs}
\usepackage{array}
\usepackage{amsmath}
\usepackage{amssymb}
\usepackage{multirow}
\usepackage{placeins}
\usepackage{tabularx}

\newcommand{\meanrss}{\mbox{MeanRSS}}
\newcommand{\mcr}{\mathrm{MCR}}
\newtheorem{proposition}{Proposition}
\newtheorem{corollary}{Corollary}
\title{Representation Redistribution in Sparse Autoencoders}

\author{Ziye Mei \quad Haotian Ma \\
{\mdseries University of Wisconsin--Madison}}

\AddToHook{env/table/begin}{\setlength{\belowcaptionskip}{4pt}}
\begin{document}
\maketitle

\begin{abstract}
Presenting the same request with different context or formatting can change which sparse autoencoder (SAE) features respond inside a language model. Understanding these changes raises two questions. Which feature responses recur across related inputs, and which of them contribute to downstream computation?

We study this phenomenon as representation redistribution and introduce the redistribution sensitivity score (RSS) to identify large feature changes that recur in a consistent direction. We select features on one set of prompt variants and evaluate them on separate variants. Across 27 pretrained SAEs spanning eight language models, RSS reveals a consistent pattern of redistribution. Context and wrapper transformations produce stronger recurring responses than local edits.

We then use two frozen safety readouts to measure how these changes contribute to downstream computation. RSS identifies features with approximately twice the mean held out contribution of features selected by the largest readout coefficients, using activation responses alone. An exact decomposition explains how feature changes and readout coefficients determine their effects on the decision margin. Combining RSS contributions with the clean margin improves recurrent failure AUROC from 0.863 to 0.935 and from 0.901 to 0.958, approaching the full readout results of 0.965 and 0.964.

These findings show that recurring changes in sparse representations provide a way to identify features with meaningful downstream contributions.
Code is available at \href{https://github.com/OpenSAE/Redistribution-Sensitivity-Score}{\texttt{\mbox{https://github.com/}\allowbreak \mbox{OpenSAE/}\allowbreak \mbox{Redistribution-Sensitivity-Score}}}.
\end{abstract}
\section{Introduction}

Sparse autoencoders (SAEs) decompose language model activations into sparse
features, providing a way to study how information is represented inside
language models \citep{elhage2022superposition,cunningham2023sparse}.
Consider a question presented with an irrelevant sentence, a format wrapper,
or a small change in wording. The underlying request remains the same, yet
the model can activate a different combination of SAE features. Previous work
has observed changes in feature responses across related inputs
\citep{chanin2025absorption,cho-hockenmaier-2025-toward,
verma-etal-2026-multilingual} and changes in model behavior under prompt
variation \citep{sclar2024format,mizrahi2024multiprompt}. These observations
motivate a closer examination of which changes recur across related inputs.

A feature can respond strongly to one prompt variant and respond differently
to another. We study the changes that recur across multiple variants and the
features that carry them. We call this shift in activity across sparse features
\emph{representation redistribution}. This raises
two connected questions. Which feature changes recur when a prompt is varied
in related ways? How do these changes contribute to downstream computation?
Together, these questions connect the behavior of sparse representations to
the computations that use them.

\begin{figure}[t!]
  \centering
  \IfFileExists{figures/figure_1_overview.pdf}{%
    \includegraphics[width=\textwidth,keepaspectratio]{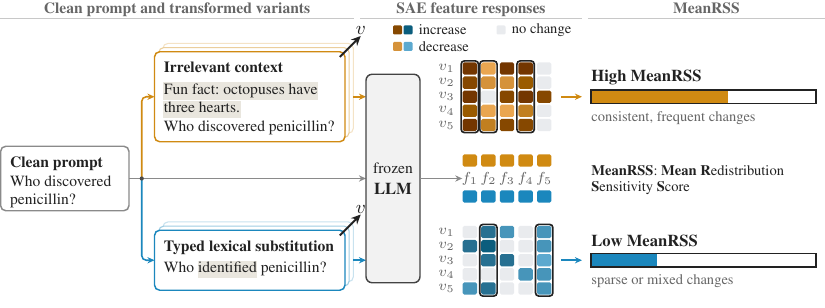}%
  }{%
    \fbox{\parbox[c][2.0in][c]{0.94\textwidth}{\centering
      \textbf{Figure 1 placeholder.}}}%
  }
  \caption{RSS identifies recurring feature changes under prompt
  transformations. In the example with five variants, highlighted features
  respond in the same direction in at least four variants. MeanRSS summarizes
  the average absolute magnitude of the selected responses.}
  \label{fig:workflow}
\end{figure}

We introduce the redistribution sensitivity score (RSS) to identify recurring
feature changes. RSS selects features with large activation responses that
repeat in a consistent direction. We first identify these features on one set
of prompt variants. We then hold the selected features and their directions
fixed and evaluate them on separate variants. This procedure measures the
persistence of redistribution across related inputs. MeanRSS summarizes the
magnitude of the selected responses (Figure~\ref{fig:workflow}).

We next examine the functional importance of these recurring changes. A
feature's contribution to a downstream computation depends on its activation
change and how that computation uses the feature. We study this relationship
with frozen linear safety readouts, where the contribution of each feature
change can be measured exactly. The readout coefficients describe how
strongly the classifier responds to each feature, while RSS identifies the
features that repeatedly change under the current prompt transformation. In
our experiments, features selected by RSS carry larger mean held out readout
contributions than features selected by the largest readout coefficients.
We also measure how the combined contributions affect the decision margin and
recurrent readout failure. Our main contributions are as follows.

\begin{itemize}
    \item We introduce RSS to identify sparse feature changes that recur in a
    consistent direction. Across 27 pretrained SAEs spanning eight language
    models, the selected responses persist on separate prompt variants.

    \item We derive an exact decomposition of feature contributions in a fixed
    readout. The decomposition links response magnitude, readout coefficients,
    and directional alignment. The clean margin connects these contributions
    to decisions.

    \item We show that features selected by RSS carry larger mean held out
    readout contributions than selection by the largest readout coefficients.
    Compact RSS supports also reconstruct substantial readout changes and
    provide information for analyzing recurrent failures.
\end{itemize}

\section{Related Work}

Quantitative explanations of internal representations have connected semantic
features to prediction scores \citep{zhang2019decision}. Sparse autoencoder
features have been evaluated through reconstruction,
sparsity, and downstream use
\citep{bricken2023monosemanticity,cunningham2023sparse,gao2024scaling,
makelov2024principled,karvonen2025saebench,kantamneni2025useful}.
Related inputs can also alter which features respond, including absorption into
more specific features and changes under prompt or surface form variation
\citep{chanin2025absorption,cho-hockenmaier-2025-toward,
verma-etal-2026-multilingual}. Feature sensitivity has also been measured as
activation reliability across semantically related texts
\citep{tian2025sensitivity}. Monosemanticity characterizes the specificity of
features to concepts, while RSS measures recurring changes in the sparse code
under a specified family of prompt transformations. RSS selects these changes
and tests their persistence on held-out variants, complementing feature-level
interpretability measurements.

Prompt sensitivity studies show that format, ordering, and wording can alter
model behavior
\citep{zhao2021calibrate,sclar2024format,mizrahi2024multiprompt,
chatterjee2024posix}, while other work studies stable representation directions
and SAE responses to optimized perturbations
\citep{dasgupta2026invariant,li2026adversarial}. We use controlled prompt
variation to characterize recurring redistribution within the sparse basis
and its relationship to downstream behavior.

Linear probes on SAE activations test whether task relevant information can be
recovered from internal representations \citep{kantamneni2025useful}, while
feature interventions can reveal context dependent downstream effects
\citep{hoang2026downstream}. Contextual attribution studies also examine
how the contribution of an input component depends on other components or
on a fixed surrounding context \citep{ling2019alphago,ma2026ssi}. Our fixed-readout analysis directly measures the contribution of features
selected from their representation responses. Holding the fitted readout fixed
makes each contribution to the prompt-induced change exact and interpretable.
For nonlinear downstream computations, joint SAE feature effects can instead
require higher order interaction terms \citep{ma2026behavioral}.

\section{Method}

Our analysis has two stages. RSS first selects recurring feature responses
from prompt transformations. We then evaluate the fixed support on held-out
variants and measure its contribution to a frozen downstream readout, including
the remaining decision margin.

\subsection{Redistribution Sensitivity Score}
\label{sec:rss-method}

RSS selects features in three steps. Response magnitude identifies candidate
features that move substantially. Recurrence finds features that respond across
multiple transformations, and directional consistency identifies changes that
follow the same direction. The final support contains the largest responses
satisfying both requirements.

All quantities are defined for one pretrained SAE, layer, base prompt, and
transformation setting. Let $x_0$ be the clean prompt and $x_v$, $v\in V$, its
transformed variants. At the aligned target position, $a_f(x)$ is the activation
of feature $f$. Define its change and average response magnitude by
\begin{align}
\delta_{v,f}
&=a_f(x_v)-a_f(x_0),
\label{eq:delta}\\
h_f
&=\frac{1}{|V|}\sum_{v\in V}|\delta_{v,f}|.
\label{eq:feature-magnitude}
\end{align}
Here $h_f$ averages absolute changes, so opposing directions do not cancel, and
the $M$ largest values form the candidate pool $C_M$. Supplementary
Section~\ref{supp:repeat-collapse} specifies operator-level averaging for
repeated realizations.

Recurrence and direction are measured from $n$ selector observations. An
observation is an individual variant change or an operator's mean change,
depending on the protocol. Let $n_f^{+}$ count observations greater than
$\tau$, and $n_f^{-}$ count observations less than $-\tau$. Define
\begin{equation}
\begin{gathered}
r_f=\frac{n_f^{+}+n_f^{-}}{n},
\qquad
c_f=\frac{\max(n_f^{+},n_f^{-})}{n_f^{+}+n_f^{-}},\\
s_f=\operatorname{sign}(n_f^{+}-n_f^{-}).
\end{gathered}
\label{eq:dominant-direction}
\end{equation}
Here $r_f$ measures how often a feature responds, $c_f$ measures how
consistently its responses follow one direction, and $s_f$ records that
direction; $c_f=0$ when no observation responds. Together, these quantities
distinguish large recurring responses from large responses that alternate in
direction.

Eligible features satisfy $r_f\geq\rho$ and $c_f\geq\kappa$. Let
$\mathcal E_M={f\in C_M:r_f\geq\rho,\ c_f\geq\kappa}$ and
$k=\min(K,|\mathcal E_M|)$. RSS retains the $k$ eligible features with largest
$h_f$ as the responsive support $S$. Since $\kappa>1/2$, tied directions are
ineligible. For nonempty $S$,
\begin{equation}
\meanrss=\frac{1}{|S|}\sum_{f\in S}h_f.
\label{eq:meanrss}
\end{equation}
RSS selects the support, whereas MeanRSS summarizes its response magnitude.
An empty support makes the summary unavailable rather than zero.

\begin{proposition}[Constrained magnitude characterization of RSS]
\label{prop:rss-constrained}
\begin{equation}
S\in
\arg\max_{\substack{T\subseteq\mathcal E_M\|T|=k}}
\sum_{f\in T}h_f.
\label{eq:rss-constrained-objective}
\end{equation}
If $|\mathcal E_M|\geq K$, the same objective is maximized over all dictionary
features satisfying the recurrence and directional consistency constraints,
up to ties.
\end{proposition}

The eligibility constraints imply a finite sample directional margin. Let
$d_f=\max(n_f^{+},n_f^{-})$ and $o_f=\min(n_f^{+},n_f^{-})$ denote the
dominant and opposing response counts.

\begin{proposition}[Finite sample directional margin]
\label{prop:directional-margin}
Every selected feature satisfies
\begin{align}
\frac{d_f}{n}
&=r_fc_f\geq\rho\kappa,
\label{eq:directional-persistence}\\
\frac{d_f-o_f}{n}
&=r_f(2c_f-1)\geq\rho(2\kappa-1).
\label{eq:directional-margin}
\end{align}
With $n=5$ and $\rho=\kappa=0.8$, at least four observations share the
dominant direction and at most one opposes it, so changing or removing one
observation cannot reverse that direction.
\end{proposition}

The selection observations establish the candidate support and its direction.
Evaluating the fixed $S$ and $s_f$ on disjoint $V_{\mathrm{ho}}$ then tests whether
these responses recur beyond the observations used to select them.

\subsection{Readout Effects of Redistribution}
\label{sec:margin-method}

A frozen linear readout provides a direct test of the functional relevance of
recurring feature responses. The coefficient magnitude $|w_f|$ measures the
readout's sensitivity to a unit activation change. The realized contribution
also depends on how much the activation changes under the current prompt
transformation. RSS identifies these changes from the representation itself,
allowing the readout analysis to measure their downstream effects independently.
Let $\widetilde a_f(x)$ denote activation scaled using clean training data.
RSS is defined on raw feature changes, so scaling for probe fitting preserves
the selected support $S$. Supplementary Section~\ref{supp:probe-training}
describes probe fitting and threshold calibration. Define
\begin{align}
\widetilde\delta_{v,f}
&=\widetilde a_f(x_v)-\widetilde a_f(x_0),
\label{eq:standardized-change}\\
z(x)
&=\mathbf w^\top\widetilde{\mathbf a}(x)+b,
\nonumber\\
m(x)
&=y[z(x)-t],
\qquad
m_0=m(x_0),
\label{eq:margin}
\end{align}
where $\mathbf w,b$ define the fixed readout,
$y\in{-1,+1}$ is the class sign, and $t$ the fixed threshold.
Positive margin denotes a correct decision, with $m_0$ its clean value.

The next question is how the selected responses affect the readout margin.
Their effects depend on response magnitude, contribution direction, and
cancellation across variants. For a variant set $U$, define average response
magnitude $A_f^U$, alignment $\eta_f^U$ with the RSS-selected direction, and
signed margin erosion $e_f^U$:
\begin{align}
A_f^U
&=\frac{1}{|U|}\sum_{v\in U}|\widetilde\delta_{v,f}|,
\label{eq:response-magnitude-standardized}\\
\eta_f^U
&=
\frac{s_f\sum_{v\in U}\widetilde\delta_{v,f}}
{\sum_{v\in U}|\widetilde\delta_{v,f}|},
\qquad -1\leq\eta_f^U\leq1,
\label{eq:directional-factor}\\
e_f^U
&=
-yw_f\frac{1}{|U|}
\sum_{v\in U}\widetilde\delta_{v,f}.
\label{eq:variant-set-contribution}
\end{align}
$A_f^U$ and $e_f^U$ apply to every readout feature, while $\eta_f^U$ is
defined relative to $s_f$ for $f\in S$. Positive $e_f^U$ erodes the margin;
the sign of $\eta_f^U$ indicates whether the net response over $U$ follows
$s_f$, while $|\eta_f^U|$ decreases with cancellation. We set $\eta_f^U=0$
when its denominator is zero. These quantities aggregate over $U$;
Section~\ref{sec:localization-method} treats individual held out variants
separately.

\begin{proposition}[RSS direction and readout contribution]
\label{prop:direction-readout}
For $f\in S$,
\begin{equation}
e_f^U=-yw_fs_fA_f^U\eta_f^U,
\qquad
|e_f^U|=|w_f|A_f^U|\eta_f^U|
\leq|w_f|A_f^U.
\label{eq:contribution-factorization}
\end{equation}
For nonzero contribution, $\operatorname{sign}(-yw_fs_f)$ predicts
$\operatorname{sign}(e_f^U)$ exactly when $\eta_f^U>0$.
\end{proposition}
Equation~\ref{eq:contribution-factorization} shows why selecting the largest
readout coefficients alone can miss the features that carry the largest
transformation-induced contributions: the actual response magnitude and its
directional alignment both matter.

On the selection variants, write $e_f=e_f^{V_{\mathrm{sel}}}$. The full and
selected contributions to average margin erosion are
\begin{equation}
E=\sum_f e_f,
\qquad
E_S=\sum_{f\in S}e_f,
\label{eq:aligned-erosion}
\end{equation}
where $E$ is full average margin erosion and $E_S$ the part carried by the RSS
support.

\subsection{Support Fidelity and Residual Margin}
\label{sec:support-task-method}

Margin erosion measures the contribution of transformed features to the
readout's score change. To determine how these contributions affect the
original decision, we combine their erosion with the clean margin:
\begin{align}
m_{\mathrm{res}}&=m_0-E,
&
m_{\mathrm{res}}^{S}&=m_0-E_S,
\label{eq:residual-margins}\\
m_{\mathrm{res}}&=
\frac{1}{|V_{\mathrm{sel}}|}
\sum_{v\in V_{\mathrm{sel}}}m(x_v).
\label{eq:residual-margin-identity}
\end{align}
Thus $m_{\mathrm{res}}$ is exactly the average transformed margin on
$V_{\mathrm{sel}}$, while $m_{\mathrm{res}}^{S}$ replaces full erosion by
$E_S$ but retains the full clean margin $m_0$.

On held out variants, let
$e_f^{\mathrm{ho}}=e_f^{V_{\mathrm{ho}}}$. For nonzero total absolute contribution, Margin Contribution Retention
(MCR) measures how much of the readout's held-out contribution is carried by
the selected support $S$:
\begin{equation}
\mcr=
\frac{\sum_{f\in S}|e_f^{\mathrm{ho}}|}
{\sum_f|e_f^{\mathrm{ho}}|}.
\label{eq:mcr}
\end{equation}
Because absolute values are taken after averaging each feature's contribution
over $V_{\mathrm{ho}}$, MCR measures retained contribution magnitude rather
than signed margin fidelity.

Signed fidelity also depends on transfer of the selected contributions and the
net contribution of omitted features. Supplementary
Section~\ref{supp:theory} gives the exact cancellation decomposition and
conditions for preserving readout decisions and average margin rankings.

\subsection{Localization, Readout Counterfactuals, and Selective Review}
\label{sec:localization-method}

Average contributions describe the support across variants. We next measure
its effect on individual held-out cases: how much of each logit change is
reconstructed, and how restoring the selected inputs changes the readout
decision. Linearity gives the exact logit decomposition and the margin after
restoring selected readout inputs to their clean values:
\begin{align}
z(x_v)-z(x_0)
&=
\sum_f w_f\,\widetilde\delta_{v,f},
\label{eq:exact-logit-shift}\\
m_{\mathrm{cf}}(x_v)
&=
m_0+y\sum_{f\notin S}w_f\widetilde\delta_{v,f}.
\label{eq:counterfactual-margin}
\end{align}
Restricting the first sum to $S$ reconstructs the support's logit shift. For an
erroneous transformed decision, $m_{\mathrm{cf}}(x_v)>0$ means that restoring
the selected readout inputs reverses the error. This counterfactual changes
only the readout computation, not the language model's activations or generated
behavior. For offline selective review, cases are ranked by increasing residual margin
computed on selection variants and evaluated against errors on held out
variants. 

\section{Experiments and Results}

\subsection{Systematic and Persistent Redistribution Across Pretrained SAEs}
\label{sec:representation-setup}
\label{sec:atlas-result}
\label{sec:calibration-transfer-result}

Our first experiment asks whether related prompt transformations produce
systematic, reproducible feature redistribution across pretrained SAEs. We use
1,000 English prompts spanning 12 content categories, fixed before SAE
inference. Six transformation settings let us compare changes to the surrounding
context or presentation with local punctuation and lexical edits. MeanRSS
measures the strength of recurring responses for each setting. We evaluate 27 pretrained SAEs spanning GPT-2, Pythia, Gemma 2 and 3, Mistral,
and Llama 3.1 across multiple layers. We normalize the six MeanRSS setting
means within each SAE to compare transformation profiles rather than absolute
response scales. Each setting uses five operators with three realizations
each; mean changes within each operator supply the selector observations, and
the evaluated token remains aligned to the preserved target span.
Supplementary Section~\ref{supp:protocols} and Table~\ref{tab:sae-specs} give
the full protocol, selector parameters, and SAE specifications.

\begin{figure}[t!]
\centering
\includegraphics[width=\textwidth,keepaspectratio]{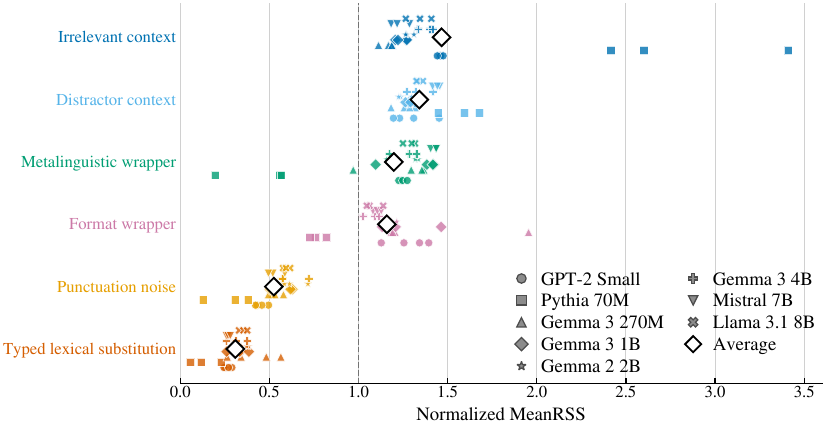}
\caption{Recurring redistribution has a systematic transformation profile.
Normalized MeanRSS compares six prompt transformations across 27 pretrained
SAEs; each SAE's values are normalized by its mean across settings. The
outlined diamonds summarize the cross-SAE averages. The prominent Pythia
profile is examined in Supplementary Section~\ref{supp:pythia-diagnostics}.}
\label{fig:rss-atlas}
\end{figure}

The normalized context and wrapper mean exceeds the local edit mean in all 27
SAEs, with an average difference of 0.874 and a 95\% interval of
$[0.867, 0.881]$ from bootstrap resampling of base prompts with the SAE panel
fixed. The pattern also holds across prompts: after aggregation
across SAEs, the context and wrapper mean exceeds the local edit mean for all
1,000 prompts, and every context or wrapper setting exceeds both local edits
for 997. Supplementary Section~\ref{supp:rss-evidence} reports support
organization and robustness to feature scaling and unavailable cases. Across four intensity levels, the median prompt level Spearman correlations
between intensity and MeanRSS are 0.95 for punctuation edits, 0.74 for
irrelevant context, and 0.50 for the format wrapper. Supports are selected
separately at each intensity, so this tests magnitude calibration rather than
persistence of the same features. Supplementary
Table~\ref{tab:intensity-levels} gives the intensity definitions.

\begin{figure}[t!]
\centering
\includegraphics[width=\textwidth,keepaspectratio]{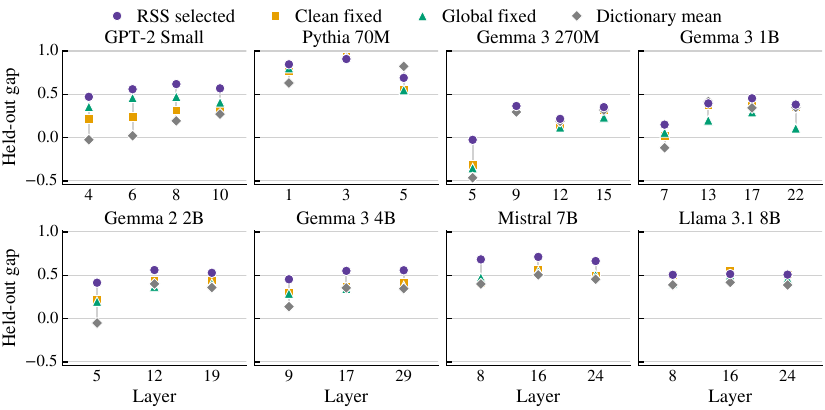}
\caption{Held-out evaluation of the transformation response contrast.
Each point measures the normalized context-and-wrapper response minus the
local-edit response for a fixed feature set and SAE layer. RSS supports retain
the contrast on variants excluded from selection. Comparisons use clean-selected
features, globally active features, and the full-dictionary mean.}
\label{fig:rss-transfer}
\end{figure}

We next test persistence beyond selection at two levels: whether a fixed support
preserves the transformation profile on held out variants, and whether its
features preserve their selected directions. Figure~\ref{fig:rss-transfer} evaluates the first question. RSS gives stronger
context and wrapper versus local edit separation than each comparison method in
24--26 of 27 SAEs and gives the largest contrast in 22. The profile also
remains stable under disjoint realization and operator holdouts
(Supplementary Section~\ref{supp:rss-evidence}). Directional consistency on the selection observations, including the finite
sample guarantee in Proposition~\ref{prop:directional-margin}, does not
guarantee the same direction on new variants. On held out variants, the
response rate in the selected direction is 0.952 for RSS and 0.758 for random
supports matched in size and candidate pool
(Supplementary Tables~\ref{tab:s-random-support}--\ref{tab:s-operator-holdout}).
Together, these experiments establish recurring redistribution as a stable,
transformation-dependent property of the selected sparse features.

\subsection{Feature Persistence and Functional Relevance in Frozen Readouts}
\label{sec:safety-setup}
\label{sec:support-task-setup}
\label{sec:support-task-result}

Our second experiment asks whether RSS-selected changes also carry meaningful
contributions to downstream computation. Frozen linear safety readouts make the
contribution of each SAE feature change exactly measurable. This setting also
lets us distinguish feature response magnitude from readout-weight importance.
The functional study uses 1,800 balanced WildGuardMix prompts
\citep{han2024wildguard}, split 1,200/300/300 into training, validation, and
test sets. Probe fitting, regularization,
threshold calibration, and layer selection use only clean training and
validation data; the resulting readouts are frozen before transformed test
evaluation. We use Gemma 3 4B layer 17 and Llama 3.1 8B layer 24, with
selection and clean performance reported in Supplementary
Table~\ref{tab:frozen-probes}. These layers were chosen from the representation
study's candidate layers using clean validation performance and the frozen
fitting protocol (Supplementary Section~\ref{supp:probe-training}).

\begin{figure}[t!]
\centering
\includegraphics[width=\textwidth,keepaspectratio]{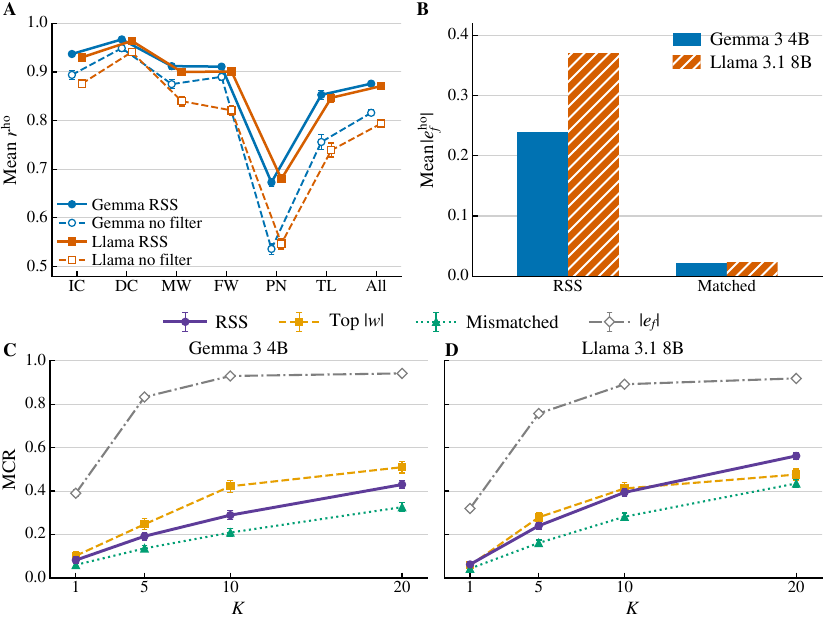}
\caption{Recurring feature responses and their functional relevance on
held-out variants. A, persistence of the direction selected by RSS compared
with magnitude-only selection. B, actual readout contribution after matching
features by coefficient magnitude. C--D, the fraction of readout contribution
retained as the support budget $K$ varies. The $|e_f|$ comparator ranks
selection-variant readout contributions.}
\label{fig:support-task-relevance}
\end{figure}

For each test prompt and transformation setting, one realization per operator
forms $V_{\mathrm{sel}}$ and the other two form disjoint $V_{\mathrm{ho}}$.
The five selection variants determine the RSS support and ranking signals,
while persistence, readout contributions, and failure are evaluated on the ten
held out variants. This tests transfer to new realizations of the same
operators without selecting on their outcomes. We define recurrent readout
failure as at least two of the ten held out margins being negative. Requiring a
correct clean prediction, available support, and complete outcomes leaves
1,524 Gemma and 1,445 Llama prompt and setting cases. Confidence intervals
use bootstrap resampling clustered by base prompt. The first test isolates the value of RSS's recurrence and directional
consistency constraints. Magnitude-only selection provides a direct comparison:
it selects large feature changes based on response magnitude alone, with the
same support size. For $f\in S$, held
out response in the selected direction is
\begin{equation}
r_f^{\mathrm{ho}}
=
\frac{1}{|V_{\mathrm{ho}}|}
\sum_{v\in V_{\mathrm{ho}}}
\mathbf{1}\!\left[s_f\,\delta_{v,f}>\tau\right].
\label{eq:heldout-recurrence}
\end{equation}
Mean held out response rates are 0.876 versus 0.816 for Gemma and 0.871 versus
0.794 for Llama (Figure~\ref{fig:support-task-relevance}A). Thus RSS identifies more reliably recurring directions than magnitude-only
selection, showing the value of its recurrence and consistency constraints.

The second test asks whether RSS identifies features that actually carry
readout change. Matching features by $|w_f|$ isolates the contribution of the
transformation-dependent response. RSS features have mean absolute held-out
contributions of 0.240 versus 0.022 for Gemma and 0.370 versus 0.022 for Llama,
with RSS larger in 96.9\% and 98.0\% of matched pairs
(Figure~\ref{fig:support-task-relevance}B). The sign rule in
Proposition~\ref{prop:direction-readout} agrees with the sign of the
contribution averaged over held out variants in 97.4\% and 97.1\% of cases.

At $K=20$, RSS retains 43.0\% and 56.2\% of absolute held out contribution,
versus 32.6\% and 43.5\% for RSS supports selected for another transformation.
Readout-aware selection by $|e_f|$ uses selection-variant contributions
and retains more than 90\% in both models
(Figure~\ref{fig:support-task-relevance}C--D). These results show that RSS identifies reproducible feature changes with
substantial downstream contributions. Direct selection by readout contribution
serves the separate objective of maximizing retained contribution for a
particular fitted probe.

We also compare RSS directly with selecting the largest readout coefficients.
RSS yields larger mean held-out feature contributions in both safety readouts:
0.240 versus 0.110 for Gemma and 0.370 versus 0.188 for Llama
(Supplementary Table~\ref{tab:s-topw-functional}). RSS thereby identifies
features more actively involved in the current transformation-induced readout
change, using only their recurring representation responses for selection.

\subsection{Frozen Readout Failure Depends on Alignment and Remaining Margin}
\label{sec:margin-result}
\label{sec:failure-result}

\begin{figure}[t!]
\centering
\includegraphics[width=\textwidth,keepaspectratio]{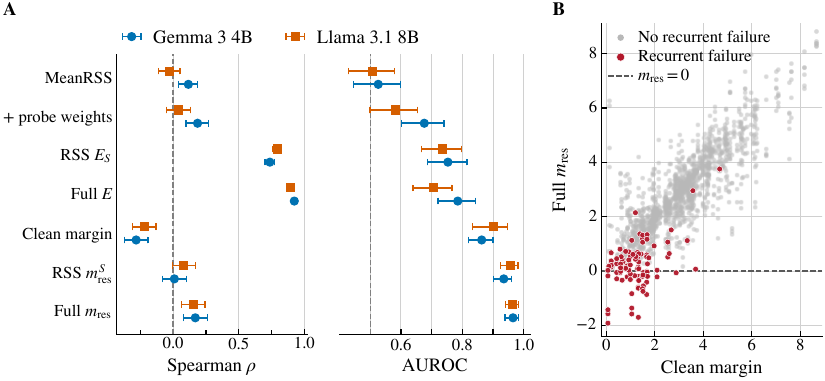}
\caption{From recurring feature changes to readout decision failure.
A, Spearman correlation measures how accurately selection-variant erosion
tracks held-out erosion, while AUROC measures the ranking of recurrent readout
failures. B, clean and full residual margins in Gemma 3 4B show how the
remaining margin relates to held-out failure; the dashed line marks zero
average selection margin.}
\label{fig:task-aligned-prediction}
\end{figure}

Our third experiment asks when recurring feature responses affect a readout
decision. A feature change first contributes a signed shift to the readout
score; its direction and the original clean margin determine whether that shift
crosses the decision threshold. We therefore compare response magnitude,
weighted magnitude, signed erosion, and residual margin with held-out erosion
and recurrent readout failure (Figure~\ref{fig:task-aligned-prediction}A).

Although probe AUROC remains above 0.90 under every transformation,
4.2--9.5\% of decisions change. MeanRSS correlates weakly with held out margin
erosion, and absolute probe weighting adds only modest information. Full
erosion estimated on selection variants correlates with held out erosion at
0.923 for Gemma and 0.896 for Llama. The corresponding RSS support
correlations are 0.737 and 0.793
(Supplementary Table~\ref{tab:s-full-support-audit}). Thus the selected
responses retain substantial information about erosion, though less than the
full readout.

Clean margin alone gives recurrent failure AUROCs of 0.863 for Gemma and 0.901
for Llama. Subtracting RSS support erosion from the full clean margin gives
AUROCs of 0.935 and 0.958, using median supports of 20 and 19 features. The
full residual margins give 0.965 and 0.964
(Figure~\ref{fig:task-aligned-prediction}; Supplementary
Table~\ref{tab:s-full-support-residual}). These are offline rankings using
known class labels, with erosion computed on selection variants and recurrent
failure measured on held out variants.

Failures concentrate at low full residual margin
(Figure~\ref{fig:task-aligned-prediction}B), and every case with negative full
residual margin has recurrent readout failure in this cohort. Because residual
margin is computed on selection variants whereas recurrent failure is defined
on held out decisions, this association provides evidence that margin effects observed during
selection transfer to new variants. The experiment establishes the roles of
signed contribution and the remaining decision margin in connecting recurring
redistribution to failure.

\subsection{Compact Supports Preserve Readout Change and Audit Utility}
\label{sec:action-result}

Our final experiment measures how effectively a compact RSS support localizes
readout change on individual held-out variants. Logit reconstruction tests how
much of the change is carried by the support, readout restoration tests its
contribution to erroneous decisions, and selective review tests the value of
these signals for ranking cases. This evaluation complements the
clean-margin-based failure rankings.

\begin{figure}[t!]
\centering
\includegraphics[width=\textwidth,keepaspectratio]{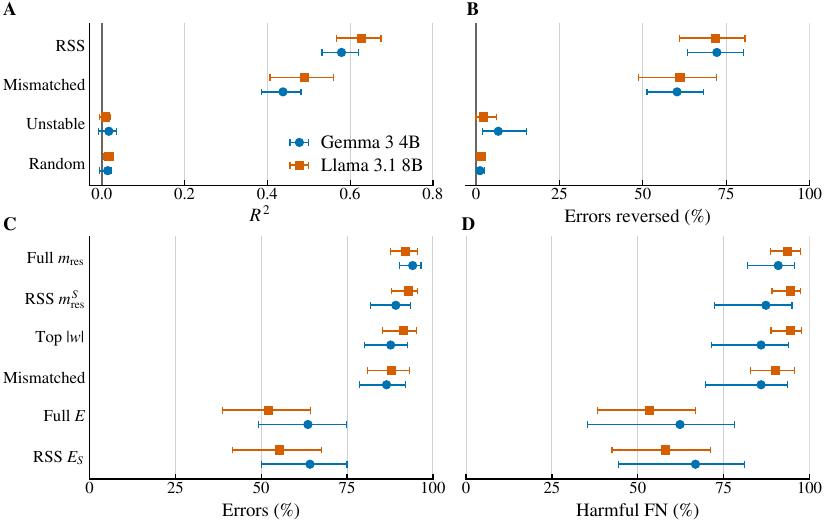}
\caption{Compact RSS supports localize readout change and support offline
auditing. A, $R^2$ measures reconstruction of held-out logit changes.
B, readout restoration measures the share of transformed probe errors reversed
by restoring the selected feature values. C--D, a 20\% offline review budget
captures probe errors and harmful false negatives. Error bars show 95\%
bootstrap confidence intervals.}
\label{fig:localization-action}
\end{figure}

RSS reconstructs held out logit changes with $R^2$ of approximately 0.58 and
0.63. Restoring selected readout inputs to their clean values reverses 72.2\%
and 71.8\% of transformed probe errors
(Figure~\ref{fig:localization-action}A--B). Mismatched supports perform worse,
whereas globally unstable and random features yield $R^2$ of only about
0.01--0.02. These comparisons localize a substantial part of the readout
change to the selected responses
(Supplementary Table~\ref{tab:localization-full}). The restoration test acts
only inside the readout.

At a 20\% offline review budget, the RSS residual margin captures 89.2\% of
Gemma errors and 92.8\% of Llama errors, together with 87.3\% and 94.4\% of
harmful false negatives (Figure~\ref{fig:localization-action}C--D). The median
supports contain 20 and 19 features, compared with 415 and 862 nonzero
coefficients in the full readouts. Supplementary
Tables~\ref{tab:s-functional-summary} and~\ref{tab:s-support-review} give the
full comparison and uncertainty intervals. The compact support therefore captures a useful portion of the transformed
readout computation and provides a concise feature-level account of its errors.
The audit combines these selected changes with the full clean readout margin.

\section{Discussion and Conclusion}

The experiments establish recurring redistribution as a measurable property of
sparse representations under controlled prompt variation. RSS identifies large,
directionally consistent feature responses that reappear on held-out variants.
The transformation profiles observed across 27 pretrained SAEs demonstrate a
structured response to changes in context and presentation, and the selected
supports provide a compact description of that response.

The functional analysis shows why these recurring responses matter. RSS
identifies features carrying substantial downstream contributions using their
representation changes alone. In the two frozen safety readouts, RSS features
carry larger realized held-out contributions than features ranked solely by
readout coefficient magnitude. Their contributions also reconstruct logit
changes and support readout-level restoration. The exact decomposition explains
how response magnitude, readout coefficient, directional alignment, and the
clean margin together determine the effects of redistribution on a decision.

These findings establish a practical connection between representation
sensitivity analysis and functional attribution. RSS supports feature-level
studies of prompt transformations and compact audits of a known downstream
computation. Our experimental evidence covers six controlled transformation
settings, their held-out realizations, and frozen linear safety readouts. This
setting enables precise measurement of individual contributions and provides a
foundation for evaluating additional downstream tasks and generation behavior.

\subsection*{AI use statement}

In this work, we used generative AI tools to improve the readability of the manuscript, to check the correctness and internal consistency of mathematical derivations and proofs, and to assist with experimental plotting.

\subsection*{Reproducibility statement}

Sections~\ref{sec:rss-method}--\ref{sec:localization-method} define RSS,
MeanRSS, readout contributions, residual margins, MCR, and the readout
counterfactual. Sections~\ref{sec:representation-setup}--\ref{sec:support-task-setup}
specify the experimental splits, comparisons, and bootstrap evaluation.
Supplementary Section~\ref{supp:theory} gives the derivations, while
Supplementary Section~\ref{supp:protocols} specifies the transformation and
alignment protocols, selector parameters, model and SAE settings, probe fitting,
and threshold calibration. Supplementary Table~\ref{tab:protocol} summarizes the
main selection and uncertainty parameters, and Supplementary
Table~\ref{tab:sae-specs} lists the pretrained SAE specifications.


\bibliography{bibliography}
\bibliographystyle{iclr2027_conference}

\clearpage
\section*{Supplementary Material}
\label{sec:supplementary}
\renewcommand{\thefigure}{S\arabic{figure}}
\renewcommand{\thetable}{S\arabic{table}}
\setcounter{figure}{0}
\setcounter{table}{0}
\setcounter{subsection}{0}
\renewcommand{\thesubsection}{S\arabic{subsection}}
\renewcommand{\thesubsubsection}{\thesubsection.\arabic{subsubsection}}

The supplementary material provides the proofs, definitions of plotted
quantities, complete protocols, and extended representation and readout results.
Table~\ref{tab:notation} summarizes the notation used throughout these analyses.

\begin{table}[!htbp]
\centering
\small
\caption{Core notation. Quantities are defined for one pretrained SAE, layer, base prompt, and transformation setting unless stated otherwise.}
\label{tab:notation}
\begin{tabularx}{\textwidth}{@{}p{0.18\textwidth}X@{}}
\toprule
Symbol & Meaning \\
\midrule
$x_0$, $x_v$ & clean prompt and its $v$th transformed variant \\
$\delta_{v,f}$ & change in SAE feature $f$ from $x_0$ to $x_v$ \\
$h_f$ & mean absolute feature change used for RSS magnitude ranking \\
$S$ & responsive support selected by RSS \\
$\meanrss$ & average $h_f$ over the responsive support \\
$\widetilde\delta_{v,f}$ & standardized feature change used by the frozen probe \\
$A_f^U$ & mean absolute standardized response of feature $f$ on variant set $U$ \\
$\eta_f^U$ & directional factor on $U$ relative to the direction selected by RSS \\
$m(x)$, $m_0$ & signed margin and clean signed margin \\
$e_f^U$ & feature contribution to margin erosion on variant set $U$ \\
$E$, $E_S$ & full and RSS support margin erosion on $V_{\mathrm{sel}}$ \\
$m_{\mathrm{res}}$, $m_{\mathrm{res}}^S$ & full and RSS support residual margin \\
$\mcr$ & fraction of absolute held out feature contribution retained by $S$ \\
\bottomrule
\end{tabularx}
\end{table}

\subsection{Mathematical Properties and Proofs}
\label{supp:theory}

We first prove the selector properties and the readout contribution
factorization. We then separate transfer error from omitted contributions and
derive conditions for preserving readout decisions and average margin rankings.
Selector thresholds and other protocol parameters are listed in Supplementary
Table~\ref{tab:protocol}.

\subsubsection{Proof of Proposition~\ref{prop:rss-constrained}}
By construction, $\mathcal E_M$ is the set of candidate features that satisfy the recurrence and directional consistency constraints, and RSS retains the $k$ members with largest $h_f$. Suppose another $k$-feature subset $T\subseteq\mathcal E_M$ had a larger value of $\sum_{f\in T}h_f$. Then at least one feature in $T\setminus S$ would have larger $h_f$ than a feature in $S\setminus T$, contradicting the ranking rule. This proves Equation~\ref{eq:rss-constrained-objective}.

If $|\mathcal E_M|\geq K$, every feature outside $C_M$ has magnitude no larger than the candidate pool cutoff, while $C_M$ already contains at least $K$ eligible features. No eligible feature outside $C_M$ can therefore increase the constrained top-$K$ objective. Ties at the candidate pool cutoff can produce additional maximizers with the same objective value.

\subsubsection{Proof of Proposition~\ref{prop:directional-margin}}
For a selected feature,
\[
d_f+o_f=n_f^++n_f^-=nr_f
\]
and
\[
\frac{d_f}{d_f+o_f}=c_f.
\]
Therefore
\[
\frac{d_f}{n}=r_fc_f
\]
and
\[
\frac{d_f-o_f}{n}
=
\frac{(d_f+o_f)(2c_f-1)}{n}
=
r_f(2c_f-1).
\]
Using $r_f\geq\rho$ and $c_f\geq\kappa>1/2$ gives Equation~\ref{eq:directional-margin}.

For $n=5$ and $\rho=\kappa=0.8$, at least four observations respond and at least 80\% of the responding observations must share the dominant direction. Hence $d_f\geq4$ and $o_f\leq1$, so $d_f-o_f\geq3$. Arbitrarily changing one observation can reduce this difference by at most two, leaving it positive; deleting one observation can reduce it by at most one. Neither operation can reverse the dominant direction.

\subsubsection{Population Response Statistics}
The finite sample directional margin is a property of the selected
observations. Population response statistics provide a separate view of why
magnitude alone does not determine directional recurrence because variation
around a mean response also matters. Under a distribution of changes, let
$\mu_f=\mathbb E[\delta_f]$ and
$v_f=\operatorname{Var}(\delta_f)$. Jensen's inequality, the triangle inequality,
and Cauchy--Schwarz give
\begin{equation}
|\mu_f|
\leq
\mathbb E|\delta_f|
\leq
|\mu_f|+\sqrt{v_f}.
\label{eq:response-magnitude-bound}
\end{equation}
If $|\mu_f|>\tau$, Cantelli's inequality gives
\begin{equation}
\Pr\!\left[\operatorname{sign}(\mu_f)\delta_f>\tau\right]
\geq
\frac{(|\mu_f|-\tau)^2}
{v_f+(|\mu_f|-\tau)^2}.
\label{eq:directional-probability-bound}
\end{equation}
These bounds relate absolute response magnitude, variation, and the probability
of a response in the direction of the population mean. They do not supply a
population guarantee for a direction selected from finite data and do not enter
Proposition~\ref{prop:directional-margin}.

For completeness, let $a=|\mu_f|-\tau>0$ and $X=\operatorname{sign}(\mu_f)(\delta_f-\mu_f)$. Then $\mathbb EX=0$, $\operatorname{Var}(X)=v_f$, and $\operatorname{sign}(\mu_f)\delta_f\leq\tau$ implies $X\leq-a$. Cantelli's inequality gives
\[
\Pr[X\leq-a]\leq\frac{v_f}{v_f+a^2},
\]
which yields Equation~\ref{eq:directional-probability-bound}.

\subsubsection{Proof of Proposition~\ref{prop:direction-readout}}
For $A_f^U>0$, Equation~\ref{eq:variant-set-contribution} gives
\[
s_f A_f^U\eta_f^U
=
s_f
\left(
\frac{1}{|U|}\sum_{v\in U}|\widetilde\delta_{v,f}|
\right)
\frac{
s_f\sum_{v\in U}\widetilde\delta_{v,f}
}{
\sum_{v\in U}|\widetilde\delta_{v,f}|
}
=
\frac{1}{|U|}
\sum_{v\in U}\widetilde\delta_{v,f},
\]
because $s_f^2=1$. Substitution into Equation~\ref{eq:variant-set-contribution} gives
\[
e_f^U=-yw_fs_fA_f^U\eta_f^U.
\]
The triangle inequality gives
\[
|\eta_f^U|
=
\frac{
\left|\sum_{v\in U}\widetilde\delta_{v,f}\right|
}{
\sum_{v\in U}|\widetilde\delta_{v,f}|
}
\leq1,
\]
so
\[
|e_f^U|=|w_f|A_f^U|\eta_f^U|\leq|w_f|A_f^U.
\]
When the contribution is nonzero, its sign differs from $\operatorname{sign}(-yw_fs_f)$ only through the sign of $\eta_f^U$, proving the final statement. The zero-response case follows from the definitions $A_f^U=\eta_f^U=e_f^U=0$.

\subsubsection{Held Out Support Fidelity and Preservation}

The main text separates held out residual-margin error into transfer of the
selected support and signed contribution omitted from it. We first make the
role of cancellation explicit. Let
\[
E^{\mathrm{ho}}=\sum_f e_f^{\mathrm{ho}},
\qquad
E_S^{\mathrm{ho}}=\sum_{f\in S}e_f^{\mathrm{ho}},
\]
and define the omitted absolute mass and cancellation factor
\begin{equation}
L_S^{\mathrm{ho}}
=
\sum_{f\notin S}|e_f^{\mathrm{ho}}|,
\qquad
\chi_S^{\mathrm{ho}}
=
\frac{
\left|\sum_{f\notin S}e_f^{\mathrm{ho}}\right|
}{
L_S^{\mathrm{ho}}
}.
\label{eq:tail-cancellation}
\end{equation}
We set $\chi_S^{\mathrm{ho}}=0$ when
$L_S^{\mathrm{ho}}=0$. By construction,
$0\leq\chi_S^{\mathrm{ho}}\leq1$. Values near one indicate little
cancellation among omitted contributions, while values near zero indicate
strong cancellation. The following identities involving MCR apply when the
total absolute held out contribution is nonzero, as required by its definition.

\begin{proposition}[Held out support fidelity]
\label{prop:heldout-support-fidelity}
The signed contribution omitted from $S$ satisfies the exact identity
\begin{equation}
\left|E^{\mathrm{ho}}-E_S^{\mathrm{ho}}\right|
=
\chi_S^{\mathrm{ho}}
(1-\mcr)
\sum_f|e_f^{\mathrm{ho}}|.
\label{eq:signed-tail-identity}
\end{equation}
Moreover,
\begin{equation}
\left|\bar m_{\mathrm{ho}}-m_{\mathrm{res}}^S\right|
\leq
\left|E_S^{\mathrm{ho}}-E_S\right|
+
\chi_S^{\mathrm{ho}}
(1-\mcr)
\sum_f|e_f^{\mathrm{ho}}|.
\label{eq:heldout-fidelity-bound}
\end{equation}
The support transfer term can further be bounded as
\begin{equation}
\left|E_S^{\mathrm{ho}}-E_S\right|
\leq
\sum_{f\in S}|w_f|
\left(
\left|A_f^{\mathrm{ho}}-A_f^{\mathrm{sel}}\right|
+
A_f^{\mathrm{sel}}
\left|\eta_f^{\mathrm{ho}}-\eta_f^{\mathrm{sel}}\right|
\right),
\label{eq:support-transfer-bound}
\end{equation}
where $A_f^{\mathrm{sel}}=A_f^{V_{\mathrm{sel}}}$,
$A_f^{\mathrm{ho}}=A_f^{V_{\mathrm{ho}}}$ and likewise for $\eta_f$.
\end{proposition}

\paragraph{Proof.}
By definition,
\[
E^{\mathrm{ho}}-E_S^{\mathrm{ho}}
=
\sum_{f\notin S}e_f^{\mathrm{ho}},
\]
so Equation~\ref{eq:tail-cancellation} gives
\[
\left|E^{\mathrm{ho}}-E_S^{\mathrm{ho}}\right|
=
\chi_S^{\mathrm{ho}}L_S^{\mathrm{ho}}.
\]
Equation~\ref{eq:mcr} implies
\[
L_S^{\mathrm{ho}}
=
(1-\mcr)\sum_f|e_f^{\mathrm{ho}}|,
\]
which proves Equation~\ref{eq:signed-tail-identity}. Since
$\bar m_{\mathrm{ho}}=m_0-E^{\mathrm{ho}}$ and
$m_{\mathrm{res}}^S=m_0-E_S$,
\[
\bar m_{\mathrm{ho}}-m_{\mathrm{res}}^S
=
(E_S-E_S^{\mathrm{ho}})
+
(E_S^{\mathrm{ho}}-E^{\mathrm{ho}}).
\]
The triangle inequality together with
Equation~\ref{eq:signed-tail-identity} proves
Equation~\ref{eq:heldout-fidelity-bound}.

For the transfer term, Proposition~\ref{prop:direction-readout} gives
\[
E_S^{\mathrm{ho}}-E_S
=
-y\sum_{f\in S}w_fs_f
\left(
A_f^{\mathrm{ho}}\eta_f^{\mathrm{ho}}
-
A_f^{\mathrm{sel}}\eta_f^{\mathrm{sel}}
\right).
\]
For each selected feature,
\[
\left|
A_f^{\mathrm{ho}}\eta_f^{\mathrm{ho}}
-
A_f^{\mathrm{sel}}\eta_f^{\mathrm{sel}}
\right|
\leq
\left|A_f^{\mathrm{ho}}-A_f^{\mathrm{sel}}\right|
+
A_f^{\mathrm{sel}}
\left|\eta_f^{\mathrm{ho}}-\eta_f^{\mathrm{sel}}\right|,
\]
because $|\eta_f^{\mathrm{ho}}|\leq1$. Summing after multiplication by
$|w_f|$ proves Equation~\ref{eq:support-transfer-bound}.

\begin{corollary}[Worst case accounting bounds]
\label{cor:worst-case-accounting}
Since $\chi_S^{\mathrm{ho}}\leq1$,
\begin{equation}
\left|E^{\mathrm{ho}}-E_S^{\mathrm{ho}}\right|
\leq
(1-\mcr)\sum_f|e_f^{\mathrm{ho}}|.
\label{eq:heldout-mcr-worstcase}
\end{equation}
On the selection variants,
\begin{equation}
\left|m_{\mathrm{res}}-m_{\mathrm{res}}^S\right|
=
\left|\sum_{f\notin S}e_f\right|
\leq
\sum_{f\notin S}|e_f|.
\label{eq:selection-accounting-worstcase}
\end{equation}
\end{corollary}

MCR controls omitted absolute contribution, whereas signed fidelity also
depends on cancellation. Setting the cancellation factor to one gives the worst
case bound, which does not use any cancellation among omitted features.

Preserving an average margin does not by itself preserve every individual
decision. For a held out variant $v$, compare the omitted signed contribution
with the margin reconstructed from selected changes. Define
\begin{equation}
\widehat m_v^S
=
m_0+
y\sum_{f\in S}w_f\widetilde\delta_{v,f}.
\label{eq:support-variant-margin}
\end{equation}

\begin{corollary}[Readout decision preservation]
\label{cor:decision-preservation}
If
\begin{equation}
\left|
\sum_{f\notin S}
yw_f\widetilde\delta_{v,f}
\right|
<
|\widehat m_v^S|,
\label{eq:decision-preservation-condition}
\end{equation}
then
\[
\operatorname{sign}m(x_v)
=
\operatorname{sign}\widehat m_v^S.
\]
\end{corollary}

\paragraph{Proof.}
Linearity gives
\[
m(x_v)
=
\widehat m_v^S
+
y\sum_{f\notin S}w_f\widetilde\delta_{v,f}.
\]
Under Equation~\ref{eq:decision-preservation-condition}, the omitted signed
contribution is too small to cross zero, so the two margins have the same
sign.

A sufficient condition for preserving an ordering between cases is that their
estimated margins differ by more than their possible transfer and omission
errors. For each case $i$, define the held out fidelity radius
\begin{equation}
B_i
=
\left|E_{S_i}^{\mathrm{ho}}-E_{S_i}\right|
+
\chi_{S_i}^{\mathrm{ho}}
(1-\mcr_i)
\sum_f|e_{i,f}^{\mathrm{ho}}|.
\label{eq:case-fidelity-radius}
\end{equation}
Proposition~\ref{prop:heldout-support-fidelity} gives
$|\bar m_{\mathrm{ho},i}-m_{\mathrm{res},i}^{S_i}|\leq B_i$.

\begin{corollary}[Held out ranking preservation]
\label{cor:heldout-ranking}
For cases $i$ and $j$,
\begin{equation}
m_{\mathrm{res},j}^{S_j}
-
m_{\mathrm{res},i}^{S_i}
>
B_i+B_j
\quad\Longrightarrow\quad
\bar m_{\mathrm{ho},j}>
\bar m_{\mathrm{ho},i}.
\label{eq:heldout-ranking-condition}
\end{equation}
\end{corollary}

\paragraph{Proof.}
The fidelity radii imply
\[
\bar m_{\mathrm{ho},i}
\leq
m_{\mathrm{res},i}^{S_i}+B_i,
\qquad
\bar m_{\mathrm{ho},j}
\geq
m_{\mathrm{res},j}^{S_j}-B_j.
\]
Under Equation~\ref{eq:heldout-ranking-condition}, the second lower bound
exceeds the first upper bound, proving the result. This condition preserves the
ordering of average held out margins, not the number of individual variants
with negative margin.

\subsubsection{Dilution Across the Full Dictionary}
Suppose $k$ of $F$ dictionary features respond with average magnitude $a$, while the remaining features have negligible change. The mean over the full dictionary is approximately $(k/F)a$. When $k\ll F$, averaging over the full dictionary reduces the response by the factor $k/F$, while averaging over the responsive support preserves the scale $a$.

\subsubsection{Exact Readout Decomposition}
For standardized activations,
\begin{equation}
z(x_v)-z(x_0)
=\mathbf w^\top\left[
\widetilde{\mathbf a}(x_v)-\widetilde{\mathbf a}(x_0)\right]
=\sum_f w_f\,\widetilde\delta_{v,f}.
\end{equation}
Restricting the final sum to $f\in S$ gives the reconstruction used in the localization analysis. Replacing the selected transformed activations by their clean values subtracts $\sum_{f\in S}w_f\widetilde\delta_{v,f}$ from the transformed logit. Multiplying by the label sign and adding the clean margin yields Equation~\ref{eq:counterfactual-margin}. Thus the readout counterfactual reverses a transformed probe error exactly when the contribution remaining outside $S$ leaves positive signed margin. The language model itself is unchanged.
\FloatBarrier

\subsection{Definitions of Plotted Quantities}
\label{supp:figure-metrics}

This section specifies the quantities plotted in the main figures, separating
representation profiles, feature contributions, and readout outcomes. All
averages use available cases only. For Figures~\ref{fig:rss-atlas} and
\ref{fig:rss-transfer}, quantities are defined within one SAE before
aggregation across SAEs.

To compare response profiles across SAEs, normalize each setting mean by the
average across the six settings. For transformation setting $g$, let
\begin{equation}
R_g
=
\operatorname{mean}_{i}\,
\operatorname{MeanRSS}_{i,g},
\qquad
\widetilde R_g
=
\frac{R_g}{\frac{1}{6}\sum_{g'}R_{g'}} ,
\label{eq:normalized-meanrss}
\end{equation}
where $i$ indexes prompts. Figure~\ref{fig:rss-atlas} plots
$\widetilde R_g$ for each SAE. Thus the six values for each SAE have mean
one.

For the transfer comparison, evaluate each selected or comparison feature set
on held out variants before computing the transformation contrast. For a
feature set $A$, define its held out response under setting $g$ as
\begin{equation}
H_g(A)
=
\operatorname{mean}_{i}\,
\operatorname{mean}_{f\in A_{i,g}}\,
\operatorname{mean}_{v\in V_{\mathrm{ho}}}
|\delta_{v,f}|.
\label{eq:heldout-response}
\end{equation}
We normalize $H_g(A)$ across the six settings in the same way as
Equation~\ref{eq:normalized-meanrss}, giving $\widetilde H_g(A)$. Let
$\mathcal C$ denote the four context and wrapper settings and
$\mathcal L$ the two local edits. The ordinate in
Figure~\ref{fig:rss-transfer} is
\begin{equation}
\Delta(A)
=
\frac{1}{4}\sum_{g\in\mathcal C}\widetilde H_g(A)
-
\frac{1}{2}\sum_{g\in\mathcal L}\widetilde H_g(A).
\label{eq:heldout-contrast}
\end{equation}

Figure~\ref{fig:support-task-relevance}A reports the cohort mean of the
held out response rate in the selected direction, $r_f^{\mathrm{ho}}$, from
Equation~\ref{eq:heldout-recurrence}. Panel B reports the cohort mean of
$|e_f^{\mathrm{ho}}|$ for RSS features and features matched by coefficient
magnitude. Panels C--D use MCR from Equation~\ref{eq:mcr}.

To evaluate transfer of the selection variant erosion estimates, define full
held out erosion as
\begin{equation}
E^{\mathrm{ho}}
=
\sum_f e_f^{\mathrm{ho}}
=
m_0-
\frac{1}{|V_{\mathrm{ho}}|}
\sum_{v\in V_{\mathrm{ho}}}m(x_v).
\label{eq:heldout-erosion}
\end{equation}
Figure~\ref{fig:task-aligned-prediction}A reports Spearman correlation
with $E^{\mathrm{ho}}$ and AUROC for recurrent failure. For residual-margin
signals, the correlation sign is reported in the direction of increasing
risk.

Localization compares the full logit change with the sum over the evaluated
support. For each held out variant, let
\[
d=z(x_v)-z(x_0),
\qquad
\widehat d_A=\sum_{f\in A}w_f\widetilde\delta_{v,f}.
\]
Figure~\ref{fig:localization-action}A reports the standard coefficient of
determination $R^2$ of $\widehat d_A$ against $d$ over the evaluated held
out rows. Panel B reports the fraction of transformed probe errors
($m(x_v)<0$) for which the corresponding readout-counterfactual margin
becomes positive. Panels C--D report the fractions of all held out probe
errors and harmful false negatives, respectively, contained in the
highest-risk 20\% of cases under each ranking signal.

\subsection{Frozen Protocols and Reproducibility}
\label{supp:protocols}

The protocols specify the input transformations and target alignment before
SAE analysis, then separate clean probe fitting from transformed evaluation.
Table~\ref{tab:protocol} lists the selection and uncertainty parameters.

\begin{table}[!htbp]
\centering
\small
\caption{Selection and uncertainty parameters. The main representation
protocol has five operator observations derived from fifteen variants per
setting. The readout audit selects from five variants and evaluates on ten
excluded variants.}
\label{tab:protocol}
\begin{tabular}{@{}lc@{}}
\toprule
Parameter & Value \\
\midrule
Selector observations, $n$ & 5 \\
Response rate threshold, $\rho$ & 0.8 \\
Directional consistency threshold, $\kappa$ & 0.8 \\
Response threshold, $\tau$ & $10^{-5}$ \\
Candidate features, $M$ & 400 \\
Support cap, $K$ & 20 \\
Target token policy & final non special in target span \\
Bootstrap resamples & 5,000 \\
\bottomrule
\end{tabular}
\end{table}

\subsubsection{Representation Corpus Protocol}
In the separate five-variant analysis, all 90,000 expected prompt, SAE, and
setting cases across the smaller four models are evaluated. Eighteen cases
(0.02\%), all under the two local edit settings, have no eligible feature and
remain unavailable rather than assigned zero. This case count is distinct from
the number of transformed input texts in the main protocol with 15 variants.

Table~\ref{tab:sae-specs} lists the checkpoint and architecture details for all 27 pretrained SAEs. The GPT-2 and Pythia SAEs use released checkpoints from \citet{bloom2024gpt2saes,tigges2024pythiasae}; the Gemma 2 and Gemma 3 checkpoints are from Gemma Scope \citep{lieberum2024gemmascope} and Gemma Scope 2 \citep{mcdougall2025gemmascope2}, respectively; the Mistral checkpoints use the residual SAEs from \citet{engels2025notall}, and the Llama checkpoints are from Llama Scope \citep{he2024llamascope}.

\begin{table}[!htbp]
\centering

\small
\caption{Pretrained SAE specifications. ``n.r.'' means not reported.}
\label{tab:sae-specs}
\setlength{\tabcolsep}{2.0pt}\renewcommand{\arraystretch}{1.12}
\begin{tabularx}{\textwidth}{@{}>{\raggedright\arraybackslash}p{0.14\textwidth} >{\raggedright\arraybackslash}p{0.075\textwidth} >{\raggedright\arraybackslash}p{0.25\textwidth} r >{\raggedright\arraybackslash}p{0.14\textwidth} >{\raggedright\arraybackslash}X@{}}
\toprule
Model & Layers & SAE checkpoint & Width & Hook & Architecture / sparsity \\
\midrule
GPT-2 Small & 4, 6, 8, 10 & \nolinkurl{GPT2-Small-SAEs-Reformatted} & 24,576 & \texttt{resid\_pre} & ReLU, $32\times$; L1 coef. $8\times10^{-5}$ \\
Pythia-70M & 1, 3, 5 & \nolinkurl{pythia-70m-deduped__res-sm} & 32,768 & \texttt{resid\_post} & ReLU, standard; sparsity n.r. \\
Gemma 3 270M & 5, 9, 12, 15 & \nolinkurl{gemma-scope-2-270m-pt} & 16,384 & \texttt{resid\_post} & JumpReLU; $L_0=18$ at layer 5 and 20 elsewhere \\
Gemma 3 1B & 7, 13, 17, 22 & \nolinkurl{gemma-scope-2-1b-pt} & 16,384 & \texttt{resid\_post} & JumpReLU; $L_0=18$ at layer 7 and 20 elsewhere \\
Gemma 3 4B & 9, 17, 29 & \nolinkurl{gemma-scope-2-4b-pt} & 65,536 & \texttt{resid\_post} & JumpReLU; nominal $L_0=53,60,60$ \\
Mistral 7B & 8, 16, 24 & \nolinkurl{mistral_7b_layer_L} & 65,536 & \texttt{resid\_pre} & ReLU, $16\times$; measured $L_0=92,74,75$ \\
Llama 3.1 8B & 8, 16, 24 & \texttt{Llama3.1-}\allowbreak\texttt{L\{8,16,24\}R-8x} & 32,768 & \texttt{resid\_post} & JumpReLU, $8\times$; measured $L_0=50$ \\
Gemma 2 2B & 5, 12, 19 & \nolinkurl{gemma-scope-2b-pt-res} \nolinkurl{width_65k} & 65,536 & \texttt{resid\_post} & JumpReLU; measured $L_0=53,72,63$ \\
\bottomrule
\end{tabularx}

\end{table}

\subsubsection{SAE Specifications}
The four larger base models, which contribute twelve of these SAEs, use bfloat16 precision without quantization, while their SAEs use float32. Checkpoint preprocessing is unchanged. Gemma 3 4B and Gemma 2 2B use none, Mistral uses per token constant norm rescaling, and Llama Scope uses checkpoint specific rescaling using dataset statistics. These settings follow the released checkpoints and are not used as a common normalization across models.

\subsubsection{Prompt Provenance and Transformations}
\label{supp:prompt-provenance}
The 1,000 clean prompts were generated before SAE inference using the GPT-5.4
snapshot \nolinkurl{gpt-5.4-2026-03-05} at temperature 0.6, in batches of
10 per content category across 12 categories. The structured generator output
included ratings for naturalness, plausibility, and category match. Acceptance
required 5--30 words and a score of at least 4 for each of these three
generator-provided ratings, followed by exact, normalized, fingerprint, and
near-duplicate filtering (token Jaccard at least 0.92) against prior generated
prompts. The recorded process accepted 1,000 prompts and rejected 108
candidates, including 106 duplicates.

Irrelevant context prepends topically unrelated benign material. Distractor
context prepends a benign competing task while retaining the original request
as operative. Metalinguistic wrappers alter response mode or presentation.
Format wrappers place the request in structured envelopes. Punctuation noise
inserts local punctuation while preserving the lexical sequence. Typed lexical
substitution replaces noncritical words from a fixed type-compatible pool.
The preserved target phrase appears exactly once in every valid variant.

Table~\ref{tab:s-prompt-examples} illustrates the effect of these operators
on one recorded prompt. The examples make the contrast between surrounding
context, presentation changes, and local edits concrete.
\begin{table}[!htbp]
\centering
\small
\caption{Examples of the transformations applied to one clean prompt.}
\label{tab:s-prompt-examples}
\begin{tabularx}{\textwidth}{@{}>{\raggedright\arraybackslash}p{0.26\textwidth}X@{}}
\toprule
Setting & Example text \\
\midrule
Clean & Maya bought two tickets for the jazz concert at the downtown theater. \\
Irrelevant context & The office window was closed before lunch. Maya bought two tickets for the jazz concert at the downtown theater. \\
Distractor context & Ignore the unrelated note that the answer may involve bicycles. Maya bought two tickets for the jazz concert at the downtown theater. \\
Metalinguistic wrapper & Restate and answer this same request: Maya bought two tickets for the jazz concert at the downtown theater. \\
Format wrapper & Prompt: Maya bought two tickets for the jazz concert at the downtown theater. \\
Punctuation noise & Maya bought two tickets, for the jazz concert at the downtown theater. \\
Typed lexical substitution & Olivia bought two tickets for the jazz concert at the downtown theater. \\
\bottomrule
\end{tabularx}
\end{table}

\subsubsection{Target Token Alignment}
Every prompt records the character span of the preserved target. For each clean prompt and variant, the evaluated position is the final non special token overlapping that span. Prefixes and wrappers can shift the token index, but the evaluated surface span remains fixed. No mapping failures are observed across all 90,000 cases in the five-variant analysis.

\subsubsection{Handling Repeated Realizations}
\label{supp:repeat-collapse}
Section~\ref{sec:rss-method} gives the generic formulation. In the analysis using 15 variants, each case contains five transformation operators with three realizations per operator. Exact duplicate texts are collapsed within each operator before selection. Let $U_o$ be the set of unique texts for operator $o$ and $\delta_{t,f}$ the feature change for text $t$. The mean change for operator $o$ is
\begin{equation}
\bar\delta_{o,f}=\frac{1}{|U_o|}\sum_{t\in U_o}\delta_{t,f},
\end{equation}
and the candidate magnitude is
\begin{equation}
h_f=\frac{1}{5}\sum_o\left[\frac{1}{|U_o|}\sum_{t\in U_o}|\delta_{t,f}|\right].
\end{equation}
An operator responds when $|\bar\delta_{o,f}|>10^{-5}$. A feature is eligible when it responds in at least four of five operators, the dominant direction share is at least 0.8, and it lies in the top $M=400$ by $h_f$. MeanRSS averages the absolute changes over operators and the selected support,
\begin{equation}
\operatorname{MeanRSS}=
\frac{1}{5|S|}\sum_o\sum_{f\in S}
\left[\frac{1}{|U_o|}\sum_{t\in U_o}|\delta_{t,f}|\right],
\end{equation}
so that repeated realizations of the same text are collapsed before averaging. This expression reduces to Equation~\ref{eq:meanrss} when every operator has the same number of unique realizations. If duplicate removal leaves different numbers of unique realizations, the expression keeps the five operators equally weighted rather than weighting them by the number of remaining texts. The table of selected supports is available for 161,996 of 162,000 cases; unavailable cases remain missing.

\subsubsection{Safety Corpus and Probe Training}
\label{supp:probe-training}
The safety corpus contains 900 harmful and 900 benign WildGuardMix prompts. Exact and near duplicates are grouped before the balanced split, and every prompt's transformed variants remain in the same split. For each feature, the scale estimated from clean training data is
\begin{equation}
\sigma_f=
\sqrt{\frac{1}{N_{\mathrm{tr}}}\sum_{i=1}^{N_{\mathrm{tr}}}a_f(x_i)^2+10^{-12}},
\end{equation}
and features with $\sigma_f<10^{-8}$ are removed.
The retained activations are standardized as
$\widetilde a_f(x)=a_f(x)/\sigma_f$. These standardized clean training
activations feed an elastic-net logistic probe with objective
\begin{equation}
\frac{1}{N}\sum_{i=1}^{N}\log\!\left(1+\exp[-y_i(\mathbf w^\top\widetilde{\mathbf a}_i+b)]\right)
+
\lambda\left[0.9\|\mathbf w\|_1+\frac{0.1}{2}\|\mathbf w\|_2^2\right].
\end{equation}
The intercept is unpenalized. Clean validation AUROC selects $\lambda$, with a sparsity tie break inside a tolerance of 0.002. A threshold calibrated on clean benign validation logits targets at most 5\% empirical false-positive rate. No transformed or test outcome enters unit, regularization, or threshold selection.

\FloatBarrier
\subsection{Extended RSS Atlas and Calibration}
\label{supp:rss-evidence}

\subsubsection{Profiles Across Models and Layers}
\begin{figure}[!htbp]
  \centering
  \includegraphics[width=\textwidth,keepaspectratio]{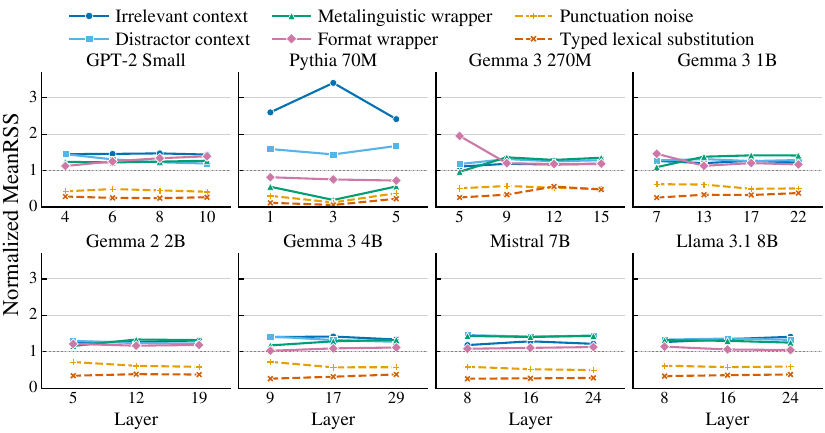}
  \caption{Normalized MeanRSS profiles for the six transformations across all sampled layers of the eight models. Normalization follows Figure~\ref{fig:rss-atlas}.}
  \label{fig:s-model layer}
\end{figure}

Figure~\ref{fig:s-model layer} resolves the aggregate transformation profile
by model and layer. Layerwise quantitative analysis has also been used to compare information
processing across network depths \citep{ma2022discarding,shwartz2017opening}.
Here we test whether the context and wrapper versus local edit separation is
shared across the sampled SAEs rather than driven by a subset of layers.

\subsubsection{Interpreting the Pythia Profile}
\label{supp:pythia-diagnostics}
Pythia shows a particularly strong response to sentence-like added context and
a comparatively small response to short instruction wrappers. In raw units,
Pythia layer 3 has MeanRSS 8.32 for irrelevant context in a five-variant
diagnostic, compared with 0.46 for
the metalinguistic wrapper and 0.32 for the format wrapper. Adding a BOS token
preserves the transformation profile. Blocking attention to inserted context
reduces the irrelevant and distractor responses to 5.5\% and 9.6\% of their
original values (median across Pythia layers). These diagnostics identify
attention to added context as a major source of the observed response.

\subsubsection{Support Organization}
\begin{figure}[!htbp]
  \centering
  \includegraphics[width=\textwidth,keepaspectratio]{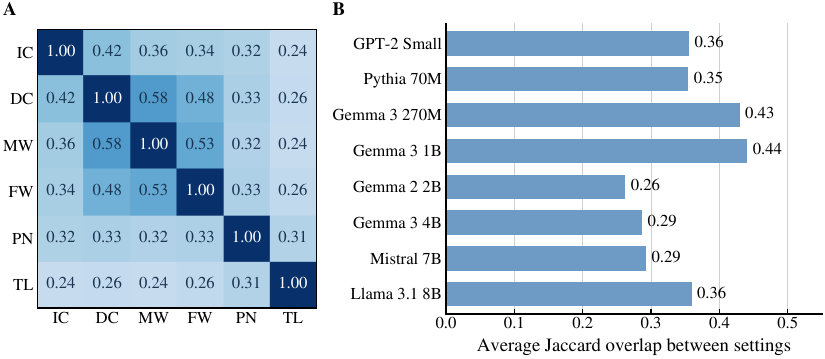}
  \caption{Overlap between the feature supports selected under different transformations, averaged over all units and 1{,}000 prompts. A, pairwise Jaccard overlap between transformations; B, averages within each model. IC denotes irrelevant context, DC distractor context, MW metalinguistic wrapper, FW format wrapper, PN punctuation noise, and TL typed lexical substitution.}
  \label{fig:s-support overlap}
\end{figure}

Similar response magnitudes need not involve the same features.
Figure~\ref{fig:s-support overlap} therefore compares selected feature
identities across transformation settings using Jaccard overlap.

\subsubsection{Support Availability and Active Feature Changes}
Tables~\ref{tab:s-support-availability} and~\ref{tab:s-active-set-change} report support availability and the changes in active SAE coordinates between clean and transformed prompts. Availability is nearly complete across all settings. Context and wrapper transformations also change the active feature set more strongly than the two local edits.

\begin{table}[!htbp]
\centering
\small
\caption{Support availability and size across transformation settings. Each setting contains 27,000 expected groups.}
\label{tab:s-support-availability}
\begin{tabular*}{\textwidth}{@{\extracolsep{\fill}}lcccc@{}}
\toprule
Setting & Available & Mean $|S|$ & Median $|S|$ & $|S|=20$ \\
\midrule
Irrelevant context & 27,000 & 19.36 & 20 & 90.3\% \\
Distractor context & 27,000 & 18.77 & 20 & 82.1\% \\
Metalinguistic wrapper & 27,000 & 19.06 & 20 & 86.6\% \\
Format wrapper & 27,000 & 18.56 & 20 & 82.1\% \\
Punctuation noise & 27,000 & 18.01 & 20 & 71.1\% \\
Typed lexical substitution & 26,996 & 16.18 & 20 & 54.6\% \\
\bottomrule
\end{tabular*}
\end{table}

Table~\ref{tab:s-unavailable-treatment} shows that the treatment of the four unavailable groups does not change the aggregate comparison.

\begin{table}[!htbp]
\centering
\small
\caption{Sensitivity to the treatment of unavailable groups. Gap is the context and wrapper mean minus the local-edit mean; brackets give 95\% confidence intervals.}
\label{tab:s-unavailable-treatment}
\begin{tabular*}{\textwidth}{@{\extracolsep{\fill}}lccc@{}}
\toprule
Treatment & Gap & 95\% CI & Order \\
\midrule
Omitted & 0.8740 & [0.8671, 0.8810] & Reference \\
Assigned zero & 0.8741 & [0.8673, 0.8810] & Unchanged \\
\bottomrule
\end{tabular*}
\end{table}

For this analysis an active coordinate is a nonzero entry in the frozen sparse
SAE encoding. For a clean prompt and one transformed prompt, let $A_0$ and
$A_v$ denote their sets of active coordinates. We report
$|A_0\cap A_v|/|A_0\cup A_v|$ for overlap,
$|A_v\setminus A_0|/|A_v|$ for the newly active fraction, and
$|A_0\setminus A_v|/|A_0|$ for the inactive fraction. When both sets are empty, overlap is undefined. When the transformed set is
empty, the newly active fraction is undefined, and when the clean set is empty,
the inactive fraction is undefined.
Within each operator, exact repeated texts are collapsed before averaging;
unique realizations are equally weighted and the five operators are then
equally weighted. Table~\ref{tab:s-active-set-change} reports these quantities
together with the positive and negative shares of absolute activation change
restricted to the RSS support. The two shares partition the support-level
absolute change whenever that change is nonzero.

\begin{table}[!htbp]
\centering
\small
\caption{Changes in active SAE coordinates between clean and transformed prompts. Overlap is Jaccard similarity, newly active is normalized by the transformed active set, and inactive is normalized by the clean active set. Positive and negative shares partition the absolute activation change within the RSS support.}
\label{tab:s-active-set-change}
\setlength{\tabcolsep}{3.2pt}
\begin{tabular*}{\textwidth}{@{\extracolsep{\fill}}lccccc@{}}
\toprule
Setting & Overlap & Newly active & Inactive & Positive & Negative \\
\midrule
Irrelevant context & 0.383 & 0.462 & 0.469 & 0.368 & 0.632 \\
Distractor context & 0.361 & 0.490 & 0.487 & 0.304 & 0.696 \\
Metalinguistic wrapper & 0.394 & 0.453 & 0.451 & 0.375 & 0.625 \\
Format wrapper & 0.381 & 0.496 & 0.478 & 0.230 & 0.770 \\
Punctuation noise & 0.692 & 0.198 & 0.187 & 0.406 & 0.594 \\
Typed lexical substitution & 0.776 & 0.133 & 0.133 & 0.381 & 0.619 \\
\bottomrule
\end{tabular*}
\end{table}

An alternative selector applies the rule in Section~\ref{sec:rss-method} directly to all 15 variants without grouping them by operator. It produces mean off-diagonal Jaccard 0.316, compared with 0.354 for the main selector. Across the same SAE and transformation combinations, the two selectors have a Spearman correlation of 0.986 for normalized MeanRSS and a Jaccard overlap of 0.698 for supports selected under the same transformation. The transformation profiles are therefore nearly identical even though the exact selected features differ.

\subsubsection{Content Categories and Model Composition}
\begin{figure}[!htbp]
  \centering
  \includegraphics[width=\textwidth,keepaspectratio]{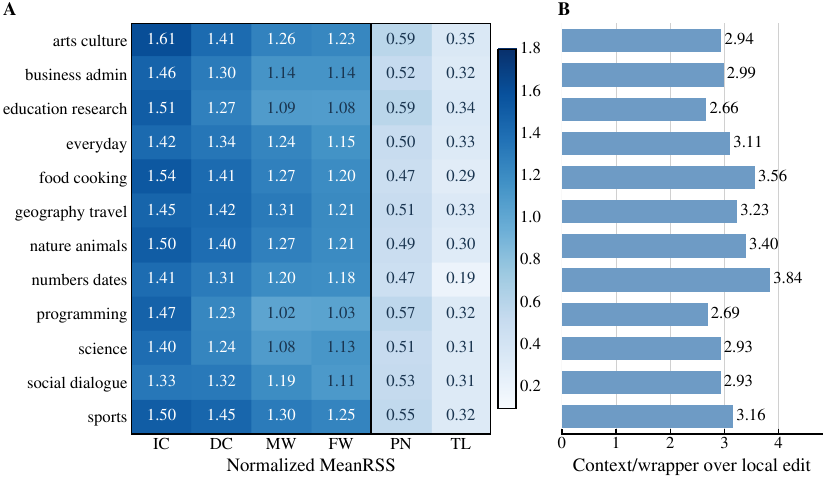}
  \caption{Normalized MeanRSS by prompt topic. A, topic level means for each transformation; B, the ratio of the context and wrapper mean to the local edit mean for each topic. Transformation abbreviations as in Figure~\ref{fig:s-support overlap}.}
  \label{fig:s-category}
\end{figure}

\begin{figure}[!htbp]
  \centering
  \includegraphics[width=\textwidth,keepaspectratio]{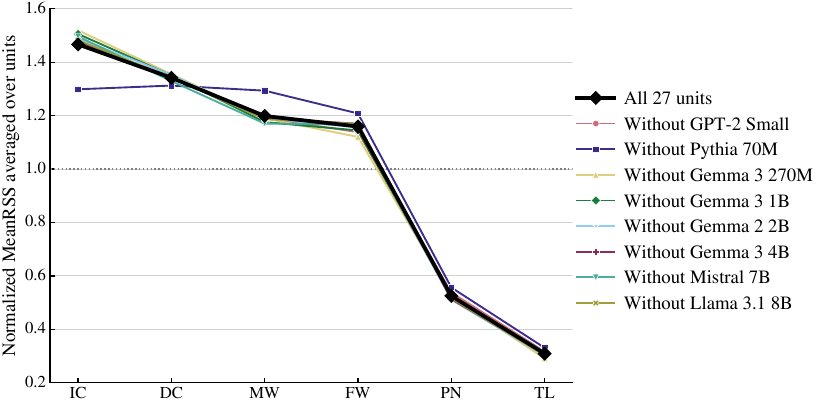}
  \caption{Profile of the six transformations after removing one model at a time, compared with the profile over all units. Transformation abbreviations as in Figure~\ref{fig:s-support overlap}.}
  \label{fig:s-family-loo}
\end{figure}

To check whether prompt composition or a particular model drives the profile,
Figure~\ref{fig:s-category} separates the results by content category, and
Figure~\ref{fig:s-family-loo} removes each model in turn.
Table~\ref{tab:primary-profile} gives the aggregate normalized profile, and
Table~\ref{tab:prompt-stability} reports consistency across prompts after
aggregation across SAEs.

\begin{table}[!htbp]
\centering
\small
\caption{Aggregate normalized MeanRSS profile with equal weight for each
SAE. Within each SAE, the six setting means are divided by their average.
Brackets give 95\% confidence intervals.}
\label{tab:primary-profile}
\begin{tabular}{@{}lc@{}}
\toprule
Setting & Normalized MeanRSS \\
\midrule
Irrelevant context & 1.466 [1.459, 1.474] \\
Distractor context & 1.341 [1.337, 1.346] \\
Metalinguistic wrapper & 1.199 [1.192, 1.205] \\
Format wrapper & 1.159 [1.154, 1.164] \\
Punctuation noise & 0.525 [0.519, 0.532] \\
Typed lexical substitution & 0.309 [0.305, 0.314] \\
\bottomrule
\end{tabular}
\end{table}

The prompt counts in Table~\ref{tab:prompt-stability} distinguish a positive
mean group difference from the stronger condition that every context or wrapper
setting exceeds both local edits.

\begin{table}[!htbp]
\centering
\small
\caption{Prompt-level consistency over the 1,000 representation-corpus prompts.}
\label{tab:prompt-stability}
\begin{tabular}{@{}lcc@{}}
\toprule
Criterion & Prompts & Share \\
\midrule
Context/wrapper mean $>$ local-edit mean & 1,000 / 1,000 & 100.0\% \\
Every context/wrapper setting $>$ both local edits & 997 / 1,000 & 99.7\% \\
\bottomrule
\end{tabular}
\end{table}

\subsubsection{Selector Diagnostics}
\begin{figure}[!htbp]
  \centering
  \includegraphics[width=0.96\textwidth,keepaspectratio]{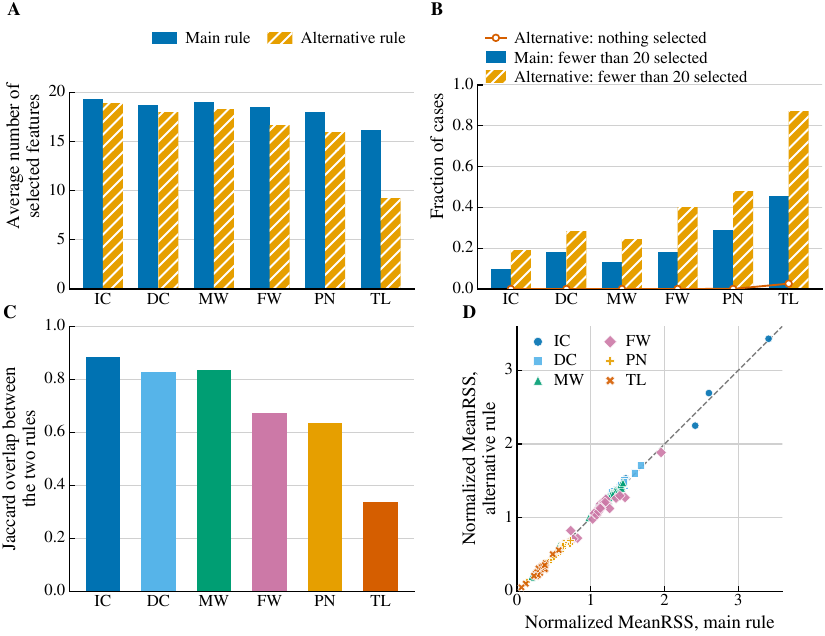}
  \caption{Comparison of the main feature selection rule with an alternative rule that treats all fifteen variants individually. A, mean support size; B, rates of undersized and empty selections; C, Jaccard overlap between the supports chosen by the two rules; D, unit level normalized MeanRSS under the two rules. Transformation abbreviations as in Figure~\ref{fig:s-support overlap}.}
  \label{fig:s-selector-diagnostics}
\end{figure}

Figure~\ref{fig:s-selector-diagnostics} compares the main selector, which uses
operator means for recurrence and direction, with selection from all fifteen
variants individually. Support sizes and overlaps show how the selected
features differ, while the normalized MeanRSS comparison tests whether those
differences change the transformation profile.

The main selector is available for 161,996 of 162,000 cases, with mean selected
count 18.32 and directional consistency 0.959 weighted by the selected
features. The alternative selector has mean selected count 16.18 and
directional consistency 0.944.

\subsubsection{Selector and Summary Robustness}
Table~\ref{tab:selector-ablation} varies selector parameters to test whether the
profile depends on the default recurrence, consistency, or support cap.
Table~\ref{tab:s-analysis-robustness} tests a different dependency by comparing
25 combinations of magnitude summaries and normalization schemes.

Feature scales can also affect magnitude ranking. We therefore repeat RSS
after scaling each feature change by its activation scale on the 1,000 clean
representation prompts. The scale is the root mean square activation on these
prompts; features below $10^{-8}$ are excluded, and retained changes are
divided by the same scale with $10^{-12}$ added inside the square root.
Table~\ref{tab:s-standardized-rss} reports the resulting profile for
$\tau\in\{0.001,0.01,0.05,0.1\}$.

\begin{table}[!htbp]
\centering
\small
\caption{Selector parameter ablation on a stratified prompt pool. The reference uses $\rho=\kappa=0.8$ and $K=20$; other rows change one parameter. Gap is the context and wrapper mean minus the local-edit mean, with 95\% confidence intervals in brackets.}
\label{tab:selector-ablation}
\setlength{\tabcolsep}{3.0pt}
\begin{tabular*}{\textwidth}{@{\extracolsep{\fill}}lccccc@{}}
\toprule
Setting & Spearman & Minimum & Gap & Mean $|S|$ & Jaccard \\
\midrule
Reference & 1.00 & 1.00 & 0.626 [0.604, 0.646] & 13.2 & 1.00 \\
$\rho=0.6$ & 1.00 & 0.94 & 0.632 [0.610, 0.652] & 13.8 & 0.90 \\
$\kappa=0.6$ & 1.00 & 0.77 & 0.595 [0.571, 0.617] & 17.4 & 0.61 \\
$K=10$ & 1.00 & 0.89 & 0.702 [0.680, 0.723] & 8.6 & 0.75 \\
$K=50$ & 1.00 & 0.94 & 0.537 [0.514, 0.558] & 17.5 & 0.89 \\
\bottomrule
\end{tabular*}
\end{table}

The robustness counts distinguish preservation of the complete six-setting
order from preservation of only its top or bottom setting. They should not be
read as identical evidence for all three comparisons.

\begin{table}[!htbp]
\centering
\small
\caption{Transformation-ranking robustness across 25 combinations of magnitude summaries and normalization schemes.}
\label{tab:s-analysis-robustness}
\begin{tabular}{@{}lc@{}}
\toprule
Criterion & Specifications \\
\midrule
Full six-setting order preserved & 17 / 25 \\
Top setting preserved & 20 / 25 \\
Bottom setting preserved & 25 / 25 \\
\bottomrule
\end{tabular}
\end{table}

\begin{table}[!htbp]
\centering
\small
\caption{RSS after scaling feature changes using clean-prompt activations. Profile Spearman is relative to the main result; Gap is the context and wrapper mean minus the local-edit mean.}
\label{tab:s-standardized-rss}
\begin{tabular*}{\textwidth}{@{\extracolsep{\fill}}ccccc@{}}
\toprule
$\tau$ & Available & Profile Spearman & Gap & Positive units \\
\midrule
0.001 & 99.99\% & 0.943 & 0.835 & 27/27 \\
0.01 & 99.94\% & 0.943 & 0.817 & 27/27 \\
0.05 & 99.65\% & 0.943 & 0.756 & 27/27 \\
0.1 & 99.40\% & 0.943 & 0.706 & 27/27 \\
\bottomrule
\end{tabular*}
\end{table}

\subsubsection{Intensity Protocol}
Table~\ref{tab:intensity-levels} defines the four levels for each intensity
series and reports the realized token and edit counts. The levels increase
transformation strength within a setting, rather than place different
transformation settings on a common numerical scale.

Support selection is performed independently at every level, so the support is selected separately for each prompt and intensity level. Rank correlations are computed within each prompt, SAE, and transformation before averaging equally across SAEs.

\begin{table}[!htbp]
\centering
\small
\caption{Intensity-level definitions and realized values. Tokens and edits are mean counts.}
\label{tab:intensity-levels}
\begin{tabular}{@{}llcc@{}}
\toprule
Setting and level & Definition & Tokens & Edits \\
\midrule
Irrelevant context 1 & 4--8 added tokens & 4.00 & 0 \\
Irrelevant context 2 & 12--16 added tokens & 12.00 & 0 \\
Irrelevant context 3 & 24--32 added tokens & 24.00 & 0 \\
Irrelevant context 4 & 48--64 added tokens & 48.00 & 0 \\
Format wrapper 1 & single label or quote & 3.24 & 10.4 \\
Format wrapper 2 & envelope with two fields & 8.92 & 28.2 \\
Format wrapper 3 & nested wrapper & 15.96 & 44.2 \\
Format wrapper 4 & nested multi-field wrapper & 28.96 & 85.4 \\
Punctuation edits 1 & exactly 1 edit & 1.00 & 1 \\
Punctuation edits 2 & exactly 2 edits & 1.99 & 2 \\
Punctuation edits 3 & exactly 4 edits & 3.99 & 4 \\
Punctuation edits 4 & exactly 8 edits & 7.96 & 8 \\
\bottomrule
\end{tabular}
\end{table}

\subsubsection{Selection--Evaluation Split Sensitivity}
We vary how the fixed total of 15 variants is divided between selection and
evaluation. This tests sensitivity to the selection and evaluation allocation,
rather than a monotone increase in total sample size.
Table~\ref{tab:reliability} reports reliability, profile stability, group sign
stability, and support overlap. A larger selection set necessarily leaves fewer
variants for evaluation.

\begin{table}[!htbp]
\centering
\small
\caption{Split-half reliability by selection size. Brackets give 95\% confidence intervals. Profile is stability against the 15-variant reference; Sign is transformation-group sign stability; Jac. is split-to-split support Jaccard.}
\label{tab:reliability}
\setlength{\tabcolsep}{2.8pt}
\begin{tabular*}{\textwidth}{@{\extracolsep{\fill}}ccccccc@{}}
\toprule
Sel. & Eval. & Reliability & Profile & Sign & CV & Jac. \\
\midrule
3 & 12 & 0.773 [0.763, 0.782] & 0.805 & 0.960 & 0.221 & 0.393 \\
5 & 10 & 0.802 [0.792, 0.810] & 0.827 & 0.964 & 0.189 & 0.539 \\
8 & 7 & 0.714 [0.703, 0.724] & 0.773 & 0.946 & 0.268 & 0.527 \\
10 & 5 & 0.681 [0.670, 0.691] & 0.765 & 0.946 & 0.279 & 0.654 \\
\bottomrule
\end{tabular*}
\end{table}

Table~\ref{tab:reliability-methods} compares selection methods at the same
selection and evaluation sizes. RSS need not maximize every measure. For
example, the dictionary mean has higher reported reliability with three
selection variants, whereas RSS has higher group sign stability.

\begin{table}[!htbp]
\centering
\small
\caption{Reliability by selection method and selection size. Column definitions follow Table~\ref{tab:reliability}.}
\label{tab:reliability-methods}
\begin{tabular}{@{}lcccc@{}}
\toprule
Method & Sel. & Reliability & Profile & Sign \\
\midrule
RSS support & 3 & 0.773 & 0.805 & 0.959 \\
RSS support & 5 & 0.802 & 0.827 & 0.964 \\
RSS support & 8 & 0.714 & 0.773 & 0.946 \\
RSS support & 10 & 0.681 & 0.765 & 0.946 \\
\midrule
Clean support & 3 & 0.772 & 0.809 & 0.921 \\
Clean support & 5 & 0.752 & 0.792 & 0.918 \\
Clean support & 8 & 0.619 & 0.709 & 0.889 \\
Clean support & 10 & 0.568 & 0.679 & 0.886 \\
\midrule
Global support & 3 & 0.744 & 0.776 & 0.922 \\
Global support & 5 & 0.761 & 0.781 & 0.928 \\
Global support & 8 & 0.621 & 0.703 & 0.887 \\
Global support & 10 & 0.591 & 0.694 & 0.892 \\
\midrule
Dictionary mean & 3 & 0.861 & 0.908 & 0.952 \\
Dictionary mean & 5 & 0.798 & 0.865 & 0.937 \\
Dictionary mean & 8 & 0.695 & 0.791 & 0.918 \\
Dictionary mean & 10 & 0.617 & 0.726 & 0.891 \\
\bottomrule
\end{tabular}
\end{table}

\subsubsection{Held Out Support Validation}
The following tests separate persistence on new realizations from persistence
on an operator omitted during selection. Table~\ref{tab:s-fixed-support-folds}
uses different realizations for selection and evaluation while keeping the
selected support fixed. Table~\ref{tab:s-random-support} tests whether this
support is more informative than random features from the same candidate pool
and of the same size. Table~\ref{tab:s-operator-holdout} then excludes an entire
operator from selection.

\begin{table}[!htbp]
\centering
\small
\caption{Fixed-support evaluation with one realization per operator used for selection. Profile is Spearman correlation with the 15-variant reference; Gap is the context and wrapper mean minus the local-edit mean; Pos. units is the number of SAEs with positive Gap; Group sign and Pairwise are prompt-level ordering rates.}
\label{tab:s-fixed-support-folds}
\setlength{\tabcolsep}{1.8pt}
\begin{tabular*}{\textwidth}{@{\extracolsep{\fill}}cccccccc@{}}
\toprule
Fold & Avail. & Profile & Gap & Pos. units & Same dir. & Group sign & Pairwise \\
\midrule
0 & 99.985\% & 1.000 & 0.909 & 27/27 & 0.940 & 99.92\% & 93.49\% \\
1 & 99.985\% & 1.000 & 0.896 & 27/27 & 0.957 & 99.86\% & 92.00\% \\
2 & 99.989\% & 1.000 & 0.897 & 27/27 & 0.959 & 99.87\% & 92.15\% \\
\bottomrule
\end{tabular*}
\end{table}

Table~\ref{tab:s-random-support} reports both absolute response and agreement
with the direction chosen during selection. The random comparison is matched
in support size and candidate pool.

\begin{table}[!htbp]
\centering
\small
\caption{Held out response for RSS and random supports matched in candidate pool and support size. Brackets give 95\% confidence intervals.}
\label{tab:s-random-support}
\begin{tabular*}{\textwidth}{@{\extracolsep{\fill}}lcccc@{}}
\toprule
Support & Absolute response & 95\% CI & Same direction & 95\% CI \\
\midrule
RSS & 107.670 & [106.749, 108.597] & 0.952 & [0.951, 0.952] \\
Random, 100 draws & 5.797 & [5.748, 5.846] & 0.758 & [0.758, 0.759] \\
\bottomrule
\end{tabular*}
\end{table}

Across the 81 unit-fold comparisons, RSS has the larger context and wrapper versus local edit difference in 76.

In Table~\ref{tab:s-operator-holdout}, the withheld operator contributes no
selection observations. Comparing requirements of four versus three responding
operators among the remaining four makes the availability and direction
persistence tradeoff explicit.

\begin{table}[!htbp]
\centering
\small
\caption{Evaluation with one transformation operator excluded from support selection. Gap is the context and wrapper mean minus the local-edit mean.}
\label{tab:s-operator-holdout}
\begin{tabular*}{\textwidth}{@{\extracolsep{\fill}}lccc@{}}
\toprule
Selection requirement & Available & Same direction & Gap \\
\midrule
4 of 4 operators & 99.42\% & 0.894 & 0.925 \\
3 of 4 operators & 100.00\% & 0.848 & 0.896 \\
\bottomrule
\end{tabular*}
\end{table}

\FloatBarrier
\subsection{Extended Audit Results}
\label{supp:monitor-evidence}

Mismatched RSS uses a support selected for another transformation setting.
For each case, comparison settings are traversed in the frozen cyclic order,
repeated feature indices are removed by first occurrence, and the resulting
support is truncated to the size of the current RSS support; the comparator is
unavailable if there are too few unique features.

\subsubsection{Probe Performance}
\begin{figure}[!htbp]
  \centering
  \includegraphics[width=0.94\textwidth,keepaspectratio]{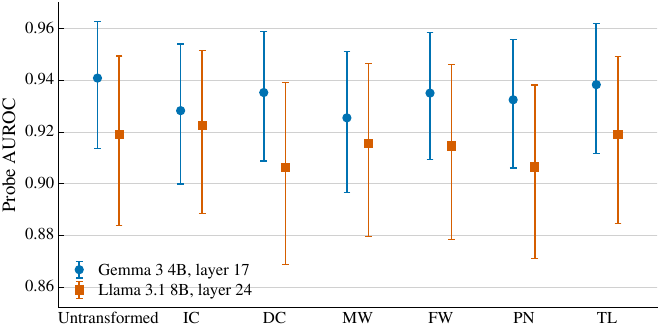}
  \caption{Probe AUROC for the two safety readouts on untransformed test prompts and under each transformation. Transformation abbreviations as in Figure~\ref{fig:s-support overlap}. Error bars show 95\% bootstrap confidence intervals.}
  \label{fig:s-clean-probes}
\end{figure}

Table~\ref{tab:frozen-probes} reports the selected probes and their clean test
performance. The candidate representation-study layers were 9, 17, 29 for
Gemma 3 4B and 8, 16, 24 for Llama 3.1 8B. Llama layer 24 gave the highest
clean validation AUROC (0.935, versus 0.918 and 0.919); Gemma layer 17 was
the eligible fit under the original frozen convergence budget. The selection
used clean training and validation outcomes. Figure~\ref{fig:s-clean-probes}
compares clean and transformed AUROC under each setting while keeping the
fitted coefficients and calibrated threshold fixed.

\begin{table}[!htbp]
\centering
\small
\caption{Safety probes and clean test performance. Brackets give 95\% confidence intervals for AUROC; FNR and FPR are false-negative and false-positive rates.}
\label{tab:frozen-probes}
\setlength{\tabcolsep}{3.0pt}
\begin{tabular*}{\textwidth}{@{\extracolsep{\fill}}lcccccc@{}}
\toprule
Model & Unit & $\lambda$ & Weights & Clean AUROC & Harmful FNR & Benign FPR \\
\midrule
Gemma 3 4B & L17 & 0.01 & 415 & 0.941 [0.914, 0.963] & 23.3\% & 7.3\% \\
Llama 3.1 8B & L24 & 0.001 & 862 & 0.919 [0.884, 0.949] & 33.3\% & 6.0\% \\
\bottomrule
\end{tabular*}
\end{table}

\subsubsection{Results by Transformation}
\begin{figure}[!htbp]
  \centering
  \includegraphics[width=0.93\textwidth,keepaspectratio]{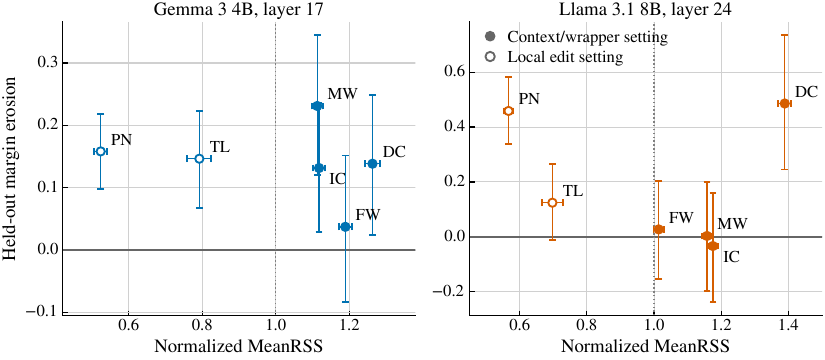}
  \caption{Normalized MeanRSS and margin erosion on held out variants for each transformation, evaluated on the audited safety units. Transformation abbreviations as in Figure~\ref{fig:s-support overlap}. Error bars show 95\% bootstrap confidence intervals.}
  \label{fig:s-setting-profiles}
\end{figure}

The transformation profile describes response magnitude, but it need not order
readout effects in the same way. Figure~\ref{fig:s-setting-profiles} and
Table~\ref{tab:safety-setting-full} compare normalized MeanRSS with held out
margin erosion, decision changes, and probe AUROC for each setting.

\begin{table}[!htbp]
\centering
\small
\caption{MeanRSS and probe outcomes by transformation setting. Brackets give 95\% confidence intervals for normalized MeanRSS and erosion. Erosion is mean held out margin loss; Flip is the fraction of predictions that differ from the clean prediction.}
\label{tab:safety-setting-full}
\setlength{\tabcolsep}{2.4pt}
\begin{tabularx}{\textwidth}{@{}l >{\raggedright\arraybackslash}X cccc@{}}
\toprule
Model & Setting & Norm. MeanRSS & Erosion & Flip & AUROC \\
\midrule
\multirow{6}{*}{Gemma 3 4B}
& Irrelevant context & 1.118 [1.102,1.134] & 0.132 [0.029,0.235] & 6.1\% & 0.928 \\
& Distractor context & 1.263 [1.244,1.284] & 0.138 [0.024,0.249] & 5.6\% & 0.935 \\
& Metalinguistic wrapper & 1.113 [1.098,1.128] & 0.231 [0.121,0.344] & 6.6\% & 0.926 \\
& Format wrapper & 1.190 [1.172,1.208] & 0.037 [-0.084,0.152] & 8.0\% & 0.935 \\
& Punctuation noise & 0.524 [0.506,0.541] & 0.158 [0.098,0.218] & 4.2\% & 0.933 \\
& Typed lexical substitution & 0.792 [0.759,0.825] & 0.146 [0.068,0.223] & 4.6\% & 0.938 \\
\midrule
\multirow{6}{*}{Llama 3.1 8B}
& Irrelevant context & 1.175 [1.160,1.190] & -0.034 [-0.239,0.161] & 7.5\% & 0.923 \\
& Distractor context & 1.389 [1.370,1.408] & 0.486 [0.245,0.734] & 9.5\% & 0.906 \\
& Metalinguistic wrapper & 1.158 [1.144,1.172] & 0.003 [-0.197,0.200] & 7.8\% & 0.915 \\
& Format wrapper & 1.014 [0.998,1.030] & 0.027 [-0.154,0.203] & 6.9\% & 0.915 \\
& Punctuation noise & 0.567 [0.553,0.581] & 0.459 [0.338,0.584] & 7.1\% & 0.907 \\
& Typed lexical substitution & 0.697 [0.666,0.729] & 0.124 [-0.013,0.266] & 5.5\% & 0.919 \\
\bottomrule
\end{tabularx}
\end{table}

For Gemma, the difference in margin erosion between the context and wrapper
transformations and the local edits is $+0.338$ $[0.236,0.443]$ for benign
prompts and $-0.374$ $[-0.498,-0.259]$ for harmful prompts. Thus the same
transformation group is associated with greater erosion for one class and
less erosion for the other after projection through the probe. A shared
representation profile does not imply a shared direction of readout effect.

\subsubsection{Intensity Analysis on the Safety Corpus}
\begin{figure}[!htbp]
  \centering
  \includegraphics[width=0.94\textwidth,keepaspectratio]{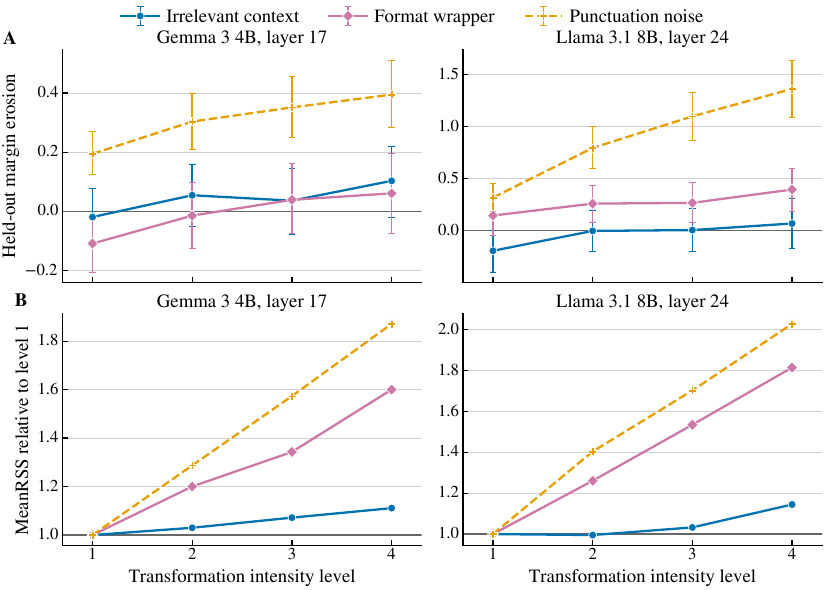}
  \caption{Effect of transformation strength on the audited safety units. A, margin erosion under three transformations rebuilt at four increasing intensity levels, with 95\% bootstrap confidence intervals. B, MeanRSS for the same series, relative to the weakest level.}
  \label{fig:s-safety-dose}
\end{figure}

The safety intensity analysis tests whether increases in representation
response are accompanied by increases in margin erosion.
Figure~\ref{fig:s-safety-dose} reports both quantities, and
Table~\ref{tab:safety-intensity-calibration} separates rank correlation from
monotonicity across all four levels.

The monotonicity denominators use only clean-correct prompts among those
eligible for the intensity analysis. For Gemma, these subsets contain 254
prompts for irrelevant context and format wrappers and 222 for punctuation
noise. For Llama, they contain 241 and 212 prompts, respectively. The numerators
count prompts for which the evaluated mean response is nondecreasing across
all four levels; they do not use the full eligible counts as denominators.

\begin{table}[!htbp]
\centering
\small
\caption{Prompt-level intensity calibration in the safety corpus. Monotonic counts use prompts whose mean response is nondecreasing across the four levels.}
\label{tab:safety-intensity-calibration}
\setlength{\tabcolsep}{2.4pt}
\begin{tabular*}{\textwidth}{@{\extracolsep{\fill}}llccccc@{}}
\toprule
Model & Setting & Eligible & MeanRSS $\rho$ & MeanRSS mono. & Erosion $\rho$ & Erosion mono. \\
\midrule
\multirow{3}{*}{Gemma 3 4B}
& Irrelevant context & 300 & 0.8 & 78/254 & 0.4 & 51/254 \\
& Format wrapper & 300 & 1.0 & 203/254 & 0.4 & 63/254 \\
& Punctuation noise & 265 & 1.0 & 137/222 & 0.4 & 37/222 \\
\midrule
\multirow{3}{*}{Llama 3.1 8B}
& Irrelevant context & 300 & 0.8 & 67/241 & 0.2 & 39/241 \\
& Format wrapper & 300 & 1.0 & 192/241 & 0.4 & 51/241 \\
& Punctuation noise & 265 & 1.0 & 138/212 & 0.6 & 62/212 \\
\bottomrule
\end{tabular*}
\end{table}

\subsubsection{Full Risk Results}
Table~\ref{tab:full-risk-results} compares risk signals on the common evaluation
cohort. Correlation with held out erosion evaluates transfer of margin change,
whereas AUROC and average precision (AP) evaluate ranking of recurrent failure.
These are distinct targets, so a signal can track erosion well without giving
the strongest failure ranking.

\begin{table}[!htbp]
\centering
\small
\caption{Risk signals on the common held out cohort. Brackets give 95\% confidence intervals. Erosion $\rho$ is Spearman correlation with held out margin erosion; AUROC and AP use recurrent failure. Residual-margin correlation is reported in the direction of increasing risk.}
\label{tab:full-risk-results}
\setlength{\tabcolsep}{2.4pt}
\begin{tabularx}{\textwidth}{@{}l >{\raggedright\arraybackslash}X ccc@{}}
\toprule
Model & Risk signal & Erosion $\rho$ & Failure AUROC & AP \\
\midrule
\multirow{9}{*}{Gemma 3 4B}
& Clean margin & -0.282 [-0.370,-0.192] & 0.863 [0.821,0.900] & 0.294 \\
& Full $\ell_1$ & 0.101 [0.025,0.173] & 0.520 [0.444,0.595] & 0.072 \\
& Full cosine & 0.083 [0.006,0.157] & 0.486 [0.398,0.572] & 0.063 \\
& MeanRSS & 0.116 [0.040,0.188] & 0.526 [0.446,0.599] & 0.069 \\
& Probe exposure & 0.104 [0.029,0.178] & 0.597 [0.510,0.683] & 0.107 \\
& MeanRSS $\times$ exposure & 0.130 [0.049,0.210] & 0.606 [0.524,0.689] & 0.091 \\
& + probe weights & 0.186 [0.097,0.272] & 0.676 [0.601,0.740] & 0.125 \\
& Margin erosion $E$ & 0.923 [0.909,0.935] & 0.785 [0.720,0.841] & 0.198 \\
& Residual margin $m_{\mathrm{res}}$ & 0.169 [0.077,0.261] & 0.965 [0.940,0.982] & 0.805 \\
\midrule
\multirow{9}{*}{Llama 3.1 8B}
& Clean margin & -0.220 [-0.307,-0.130] & 0.901 [0.833,0.947] & 0.505 \\
& Full $\ell_1$ & 0.039 [-0.046,0.123] & 0.466 [0.372,0.554] & 0.079 \\
& Full cosine & -0.024 [-0.104,0.060] & 0.528 [0.427,0.627] & 0.136 \\
& MeanRSS & -0.028 [-0.109,0.052] & 0.506 [0.429,0.578] & 0.086 \\
& Probe exposure & 0.041 [-0.029,0.110] & 0.569 [0.482,0.652] & 0.105 \\
& MeanRSS $\times$ exposure & 0.011 [-0.067,0.091] & 0.570 [0.483,0.648] & 0.109 \\
& + probe weights & 0.039 [-0.053,0.130] & 0.581 [0.499,0.654] & 0.127 \\
& Margin erosion $E$ & 0.896 [0.877,0.912] & 0.705 [0.639,0.767] & 0.191 \\
& Residual margin $m_{\mathrm{res}}$ & 0.152 [0.065,0.243] & 0.964 [0.940,0.982] & 0.850 \\
\bottomrule
\end{tabularx}
\end{table}

\subsubsection{Relevance of the Selected Features on New Variants}

Tables~\ref{tab:s-support-task-feature}--\ref{tab:s-mcr-summary} separately
assess direction persistence, contribution after matching coefficient
magnitude, and the fraction of absolute contribution retained. These
comparisons distinguish stable response selection from selection optimized
for a fixed readout.

\begin{table}[!htbp]
\centering
\small
\caption{Held out response in the direction selected by RSS compared with magnitude-only selection. Brackets give 95\% confidence intervals.}
\label{tab:s-support-task-feature}
\begin{tabular*}{\textwidth}{@{\extracolsep{\fill}}lcc@{}}
\toprule
Model & RSS & Magnitude only \\
\midrule
Gemma 3 4B & 0.876 [0.872, 0.879] & 0.816 [0.810, 0.822] \\
Llama 3.1 8B & 0.871 [0.867, 0.874] & 0.794 [0.788, 0.800] \\
\bottomrule
\end{tabular*}
\end{table}

Table~\ref{tab:s-support-contribution} compares absolute feature contributions
after matching readout coefficient magnitude. Direction agreement compares the
observed contribution sign with $\operatorname{sign}(-yw_fs_f)$, including the
class sign and readout coefficient as well as the RSS selected direction.

\begin{table}[!htbp]
\centering

\small
\caption{Held out feature contribution after matching selected and
unselected features by readout coefficient magnitude. Contribution columns
report mean absolute contribution. Direction agreement compares observed
contribution sign with $\operatorname{sign}(-yw_fs_f)$. Brackets give 95\%
confidence intervals for the paired difference.}
\label{tab:s-support-contribution}
\setlength{\tabcolsep}{3.0pt}
\begin{tabular*}{\textwidth}{@{\extracolsep{\fill}}lcccc@{}}
\toprule
Model & RSS contribution & Matched contribution & Difference & Direction agreement \\
\midrule
Gemma 3 4B & 0.2397 & 0.0221 & 0.2177 [0.2005, 0.2356] & 97.4\% \\
Llama 3.1 8B & 0.3702 & 0.0220 & 0.3483 [0.3235, 0.3739] & 97.1\% \\
\bottomrule
\end{tabular*}

\end{table}

\begin{table}[!htbp]
\centering
\small
\caption{Direct RSS versus Top-$|w|$ comparison: mean absolute held-out
readout contribution among selected features with nonzero readout coefficients.
Larger values indicate greater realized contribution under the current
transformation.}
\label{tab:s-topw-functional}
\begin{tabular}{@{}lcc@{}}
\toprule
Model & RSS & Top-$|w|$ \\
\midrule
Gemma 3 4B & 0.240 & 0.110 \\
Llama 3.1 8B & 0.370 & 0.188 \\
\bottomrule
\end{tabular}
\end{table}

Table~\ref{tab:s-mcr-summary} compares MCR at $K=20$. The $|e_f|$ ranking uses
contributions computed from selection variants, not the held out contributions
used in the reported MCR.

\begin{table}[!htbp]
\centering
\small
\caption{Margin Contribution Retention at $K=20$. Mismatched RSS uses a support selected for another transformation; $|e_f|$ ranks features by contributions computed from $V_{\mathrm{sel}}$. Brackets give 95\% confidence intervals.}
\label{tab:s-mcr-summary}
\begin{tabular*}{\textwidth}{@{\extracolsep{\fill}}lccc@{}}
\toprule
Model & RSS & Mismatched RSS & $|e_f|$ \\
\midrule
Gemma 3 4B & 0.430 [0.411, 0.450] & 0.326 [0.308, 0.346] & 0.942 [0.936, 0.948] \\
Llama 3.1 8B & 0.562 [0.547, 0.577] & 0.435 [0.417, 0.453] & 0.919 [0.912, 0.925] \\
\bottomrule
\end{tabular*}
\end{table}

Table~\ref{tab:s-mcr-summary} reports MCR on $V_{\mathrm{ho}}$. For a fixed case, all supports share the same denominator, so higher MCR means less omitted absolute contribution. Equation~\ref{eq:signed-tail-identity} shows that signed error additionally depends on cancellation through $\chi_S^{\mathrm{ho}}$.

Table~\ref{tab:s-full-support-audit} compares full and support erosion on the
common evaluation cohort. Correlation measures agreement with held out erosion;
AUROC uses erosion as a score for recurrent failure.

\begin{table}[!htbp]
\centering
\small
\caption{Margin erosion results for the full readout and RSS support.
Erosion $\rho$ is correlation with held out erosion, and Failure AUROC ranks
recurrent failure using the corresponding erosion score. Brackets give 95\%
confidence intervals.}
\label{tab:s-full-support-audit}
\begin{tabular*}{\textwidth}{@{\extracolsep{\fill}}llcc@{}}
\toprule
Model & Feature set & Erosion $\rho$ & Failure AUROC \\
\midrule
\multirow{2}{*}{Gemma 3 4B}
& Full readout & 0.923 [0.909, 0.935] & 0.785 [0.721, 0.842] \\
& RSS support & 0.737 [0.700, 0.771] & 0.753 [0.686, 0.815] \\
\midrule
\multirow{2}{*}{Llama 3.1 8B}
& Full readout & 0.896 [0.878, 0.911] & 0.705 [0.639, 0.766] \\
& RSS support & 0.793 [0.761, 0.820] & 0.735 [0.666, 0.797] \\
\bottomrule
\end{tabular*}
\end{table}

Table~\ref{tab:s-full-support-residual} adds the full clean margin to each
erosion estimate. This evaluates the information retained when selected
changes are used to estimate the remaining margin, rather than treating the
support as a replacement for the clean readout.

\begin{table}[!htbp]
\centering
\small
\caption{Residual margin results for the same cohort. Both feature sets
use the full clean margin. Correlation with held out erosion is oriented toward
increasing risk; AUROC evaluates recurrent failure. Brackets give 95\%
confidence intervals.}
\label{tab:s-full-support-residual}
\begin{tabular*}{\textwidth}{@{\extracolsep{\fill}}llcc@{}}
\toprule
Model & Feature set & Residual $\rho$ & Residual AUROC \\
\midrule
\multirow{2}{*}{Gemma 3 4B}
& Full readout & 0.169 [0.077, 0.261] & 0.965 [0.939, 0.982] \\
& RSS support & 0.009 [-0.081, 0.105] & 0.935 [0.901, 0.961] \\
\midrule
\multirow{2}{*}{Llama 3.1 8B}
& Full readout & 0.152 [0.065, 0.243] & 0.964 [0.940, 0.982] \\
& RSS support & 0.080 [-0.006, 0.171] & 0.958 [0.924, 0.981] \\
\bottomrule
\end{tabular*}
\end{table}

\begin{table}[!htbp]
\centering
\small
\caption{Summary of the full frozen readout and RSS supports selected for
each case on the common evaluation cohort. Size is the full nonzero coefficient
count or median support size. Both residual margin signals use the full clean
margin. Errors and Harmful FN are capture rates at 20\% offline review with
known labels.}
\label{tab:s-functional-summary}
\setlength{\tabcolsep}{4.8pt}
\begin{tabular}{llccccc}
\toprule
Model & Feature set & Size & Erosion $\rho$ & Residual AUROC & Errors & Harmful FN \\
\midrule
\multirow{2}{*}{Gemma 3 4B}
& Full readout & 415 & 0.923 & 0.965 & 94.1\% & 90.9\% \\
& RSS support & 20 & 0.737 & 0.935 & 89.2\% & 87.3\% \\
\midrule
\multirow{2}{*}{Llama 3.1 8B}
& Full readout & 862 & 0.896 & 0.964 & 91.9\% & 93.5\% \\
& RSS support & 19 & 0.793 & 0.958 & 92.8\% & 94.4\% \\
\bottomrule
\end{tabular}
\end{table}

\subsubsection{Additional Llama Layers}
We apply the same functional protocol to Llama 3.1 8B layers 8 and 16 to
assess whether the compact support results extend beyond the selected layer.
Table~\ref{tab:s-additional-llama-layers} reports direction persistence,
contribution retention, residual margin AUROC, localization, and restoration.
The metrics vary across layers, so persistence and retained audit utility
should not be treated as interchangeable measures.

\begin{table}[!htbp]
\centering
\small
\caption{Functional results for two additional Llama 3.1 8B layers under the main audit protocol.}
\label{tab:s-additional-llama-layers}
\begin{tabular*}{\textwidth}{@{\extracolsep{\fill}}ccccccc@{}}
\toprule
Layer & $|S|$ & Same direction & MCR & Residual AUROC & $R^2$ & Restoration \\
\midrule
8 & 20 & 0.960 & 0.628 & 0.905 & 0.814 & 0.566 \\
16 & 20 & 0.970 & 0.578 & 0.806 & 0.692 & 0.807 \\
\bottomrule
\end{tabular*}
\end{table}

\FloatBarrier
\subsection{Localization, Selective Review, and Training Details}
\label{supp:action-evidence}

\subsubsection{Localization and Readout Counterfactuals}
The localization comparison tests whether RSS selects changes relevant to the
current transformation rather than merely unstable or arbitrary features.
Table~\ref{tab:localization-full} evaluates all listed supports on the same
rows from $V_{\mathrm{ho}}$. In both models, RSS gives higher reconstruction
$R^2$ and reverses more probe errors than a support selected for another
transformation. Features that are frequently unstable across the corpus and
random features of the same size contribute little to either result.

\begin{table}[!htbp]
\centering
\small
\caption{Reconstruction and readout-counterfactual results. Baseline supports are matched to RSS support size; brackets give 95\% confidence intervals. $Q_{\mathrm{res}}$ is the share of probe logit change recovered from the support.}
\label{tab:localization-full}
\setlength{\tabcolsep}{2.6pt}
\begin{tabularx}{\textwidth}{@{}l >{\raggedright\arraybackslash}X cccc@{}}
\toprule
Model & Support & Size & $R^2$ & Errors reversed & $Q_{\mathrm{res}}$ \\
\midrule
\multirow{4}{*}{Gemma 3 4B}
& RSS support & 20 & 0.579 [0.532, 0.621] & 0.722 [0.635, 0.803] & 0.618 \\
& Mismatched setting & 20 & 0.438 [0.386, 0.481] & 0.603 [0.512, 0.683] & 0.368 \\
& Global instability & 20 & 0.017 [-0.008, 0.035] & 0.067 [0.018, 0.151] & 0.000 \\
& Random matched & 20 & 0.014 [-0.006, 0.022] & 0.012 [0.002, 0.026] & 0.002 \\
\midrule
\multirow{4}{*}{Llama 3.1 8B}
& RSS support & 19 & 0.627 [0.568, 0.675] & 0.718 [0.610, 0.807] & 0.800 \\
& Mismatched setting & 19 & 0.490 [0.406, 0.560] & 0.611 [0.487, 0.721] & 0.590 \\
& Global instability & 19 & 0.009 [-0.006, 0.020] & 0.023 [0.000, 0.062] & 0.000 \\
& Random matched & 19 & 0.017 [0.004, 0.022] & 0.015 [0.007, 0.027] & 0.002 \\
\bottomrule
\end{tabularx}
\end{table}

\subsubsection{Selective Review}
Table~\ref{tab:routing-summary} summarizes offline selective review with the Gemma probe trained only on clean prompts, using the same cohort as the risk prediction analysis. Random is the average of 100 random rankings, with the replicate average computed inside each shared prompt cluster bootstrap draw. The residual margin ranks cases in ascending order. For a review budget $c$, residual error is the error rate among cases left unreviewed after the highest-risk $\lfloor cN\rfloor$ cases are removed. The normalized partial area under the residual error curve (pAURC) is the trapezoidal area at budgets $c\in\{0,0.05,0.10,0.20,0.30,0.40\}$ divided by $0.40$. We report $\Delta\mathrm{pAURC}$ as the mean random ranking pAURC minus the pAURC of the listed ranking, so positive values indicate lower residual error than random.

\begin{table}[!htbp]
\centering
\small
\caption{Selective-review summary for the Gemma probe. Capture columns give the fraction of errors reviewed at the stated budget. Residual error is reported at a 20\% review budget. pAURC is the normalized partial area under the residual error curve from 0 to 40\% review, and $\Delta$ pAURC is the random ranking pAURC minus the listed value; brackets give 95\% confidence intervals.}
\label{tab:routing-summary}
\setlength{\tabcolsep}{2.4pt}
\begin{tabularx}{\textwidth}{@{}>{\raggedright\arraybackslash}X ccccc@{}}
\toprule
Ranking signal & Capture@10 & Capture@20 & Harmful FN@20 & Residual error & $\Delta$ pAURC \\
\midrule
Residual margin $m_{\mathrm{res}}$ & 85.1\% & 94.1\% & 90.9\% & 0.25\% & +0.0276 [0.0175,0.0385] \\
Low clean margin & 41.3\% & 68.9\% & 60.9\% & 1.30\% & +0.0186 [0.0117,0.0267] \\
+ probe weights & 32.1\% & 47.6\% & 49.5\% & 2.20\% & +0.0109 [0.0046,0.0180] \\
MeanRSS & 11.0\% & 18.8\% & 18.2\% & 3.40\% & +0.0007 [-0.0026,0.0046] \\
\bottomrule
\end{tabularx}
\end{table}

Table~\ref{tab:s-support-review} gives the corresponding capture rates at the 20\% review budget.

\begin{table}[!htbp]
\centering
\small
\caption{Selective review at a 20\% budget for the rankings in Figure~\ref{fig:localization-action}C--D. Values are capture percentages; brackets give 95\% confidence intervals.}
\label{tab:s-support-review}
\setlength{\tabcolsep}{3.0pt}
\begin{tabularx}{\textwidth}{@{}l >{\raggedright\arraybackslash}X cc@{}}
\toprule
Model & Ranking signal & Errors & Harmful FN \\
\midrule
\multirow{5}{*}{Gemma 3 4B}
& Residual margin $m_{\mathrm{res}}$ & 94.1 [90.2, 96.5] & 90.9 [81.9, 95.7] \\
& RSS residual margin $m_{\mathrm{res}}^S$ & 89.2 [81.8, 93.5] & 87.3 [72.4, 94.9] \\
& Mismatched RSS residual & 86.5 [78.7, 92.0] & 85.9 [69.8, 93.5] \\
& Margin erosion $E$ & 63.6 [49.3, 74.8] & 62.3 [35.4, 78.2] \\
& RSS margin erosion $E_S$ & 64.2 [50.1, 75.0] & 66.8 [44.4, 81.1] \\
\midrule
\multirow{5}{*}{Llama 3.1 8B}
& Residual margin $m_{\mathrm{res}}$ & 91.9 [87.6, 95.6] & 93.5 [88.7, 97.4] \\
& RSS residual margin $m_{\mathrm{res}}^S$ & 92.8 [88.0, 95.6] & 94.4 [89.1, 97.4] \\
& Mismatched RSS residual & 88.0 [80.9, 93.1] & 90.0 [82.8, 95.6] \\
& Margin erosion $E$ & 52.2 [38.7, 64.4] & 53.3 [38.3, 66.9] \\
& RSS margin erosion $E_S$ & 55.3 [41.7, 67.5] & 58.1 [42.5, 71.2] \\
\bottomrule
\end{tabularx}
\end{table}

\subsubsection{Probe Training with MeanRSS Sampling}
\label{supp:training-sampling}

The frozen readout analysis shows that large redistribution alone is not a
failure score. A separate question is whether exposing a probe to examples
with large redistribution can help during training. We compare MeanRSS
sampling with uniform sampling under the same augmentation budget, rather than
infer a training benefit from the frozen audit.

\begin{figure}[!htbp]
  \centering
  \includegraphics[width=\textwidth,keepaspectratio]{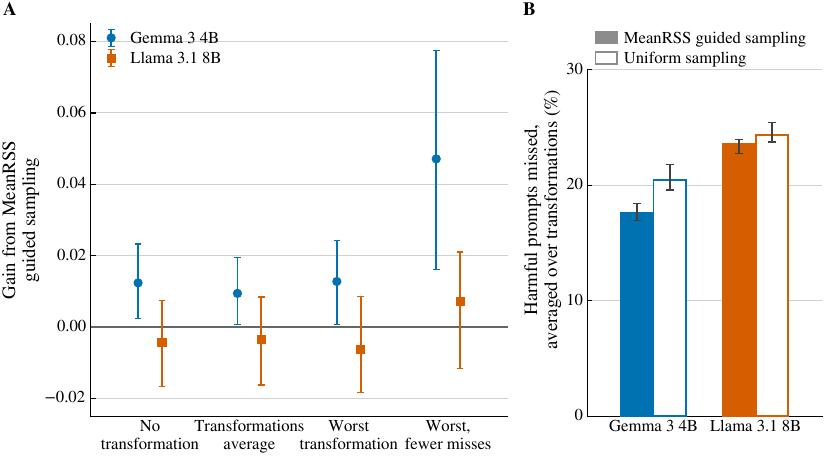}
  \caption{Augmentation guided by MeanRSS compared with uniform sampling. A, differences between the two sampling methods in clean, mean, and worst transformation AUROC, together with the reduction in harmful false negative rate under the worst transformation. Error bars show 95\% bootstrap confidence intervals. B, harmful false negative rate averaged over the six transformations across five runs. Bars show the run mean and whiskers the observed range.}
  \label{fig:training}
\end{figure}

Every augmented probe uses 1,200 clean examples and 6,000 transformed examples,
corresponding to five transformed examples per training prompt. The MeanRSS
method samples prompt and transformation pairs according to MeanRSS computed
from the training prompts, with exponent $\gamma=2$. The uniform method
samples from the same pool with equal probability. Scaling statistics and
feature masks remain those of the clean probe. Figure~\ref{fig:training} and
Table~\ref{tab:training-seed42} report the results.

For Gemma, the paired gains in the run shown in Figure~\ref{fig:training}A are
$+0.0124$ $[0.0024,0.0233]$ in clean AUROC, $+0.0094$
$[0.0007,0.0195]$ in macro AUROC, $+0.0127$ $[0.0007,0.0243]$ in worst
setting AUROC, and a reduction of $4.7$ percentage points $[1.6,7.8]$ in
harmful false-negative rate under the worst setting. Across five runs, however,
Gemma macro AUROC differences average $+0.0011$ with standard deviation
0.0051. The macro harmful false-negative rate averages 17.65\% for MeanRSS
sampling and 20.45\% for uniform sampling. The corresponding Llama rates are
23.55\% and 24.36\%.

These averages show fewer harmful prompts missed in the reported runs, but the
AUROC evidence is mixed. Table~\ref{tab:training-seed42} reports lower clean,
mean, and worst setting AUROC for Llama under MeanRSS sampling, and Gemma's
average macro AUROC gain across runs is small relative to its variation. The
training extension therefore does not establish a consistent AUROC advantage
over uniform sampling.

\begin{table}[!htbp]
\centering
\small
\caption{Retrained-probe results for Figure~\ref{fig:training}A. Clean, Mean, and Worst are AUROC; Gap is clean minus worst AUROC.}
\label{tab:training-seed42}
\setlength{\tabcolsep}{2.6pt}
\begin{tabular*}{\textwidth}{@{\extracolsep{\fill}}llccccc@{}}
\toprule
Model & Method & Clean & Mean & Worst & Worst FNR & Gap \\
\midrule
Gemma 3 4B & MeanRSS sampling & 0.945 & 0.943 & 0.938 & 0.191 & 0.0067 \\
Gemma 3 4B & Uniform & 0.933 & 0.934 & 0.926 & 0.238 & 0.0071 \\
Llama 3.1 8B & MeanRSS sampling & 0.911 & 0.910 & 0.901 & 0.256 & 0.0100 \\
Llama 3.1 8B & Uniform & 0.915 & 0.913 & 0.907 & 0.263 & 0.0082 \\
\bottomrule
\end{tabular*}
\end{table}

\FloatBarrier

\end{document}